%% file: main.tex
\PassOptionsToPackage{dvipsnames}{xcolor}
\documentclass[10pt,letterpaper]{fairmeta}
\usepackage{amsmath,amssymb,xspace}
\usepackage[shortlabels,inline]{enumitem}
\usepackage{tabularx}
\input{preamble}
\setlist[itemize]{noitemsep,leftmargin=*,topsep=3pt}
\setkeys{Gin}{width=\linewidth,keepaspectratio}

\title{Looking Right Is Not Being Right:\\Evaluating the Geometry of Agent-Built 3D Scenes}
\input{authors}
\abstract{\input{sec/0_abstract}}
\date{November 2026}
\correspondence{Ziyang Long (\email{ziyanglong@g.ucla.edu})}
\metadata[Project]{\url{https://github.com/Albertlongzi/looking-right-being-right}}
\metadata[Artifacts]{\url{https://huggingface.co/datasets/Zylong/looking-right-being-right}}
\begin{document}
\maketitle
\input{fig/main_v3_teaser}
\input{sec/1_intro}
\input{sec/2_related}
\input{sec/3_protocol}
\input{sec/5_results}

\input{sec/5b_controlled}
\input{sec/6_verification}
\input{sec/6_discussion}
\section*{Acknowledgments}
This work was conducted during Ziyang Long's internship at Amazon.
\clearpage
{\setlength{\parskip}{0pt}\setlength{\bibsep}{1pt}\def\bibfont{\fontsize{8}{10}\selectfont}\bibliographystyle{ieeenat_fullname}\bibliography{refs}}
\end{document}

%% file: preamble.tex
\usepackage{cuted}
\usepackage{currfile}
\usepackage{multirow}
\usepackage{placeins}
\usepackage{booktabs}
\usepackage{tabularx}

\newcommand{\TODO}[1]{\textcolor{red}{\textbf{TODO:} #1}}

\input{fig/data/main_v3/paper_macros}
\input{fig/data/main_v3/availability}

\input{fig/data/main_v3/figure_scope_macros}
\input{fig/data/yaw_l_audit_20261112/audit_macros}
\input{tab/main_v3_table_macro_inputs}
\input{fig/data/narrative_draft_20261112/macros}
\input{fig/working_figure_assets}
\input{fig/data/section7/section7_report_test_macros}

\input{fig/data/section7/section7_intervention_macros}

\newcommand{\XRSolCtlGain}{0.4}
\newcommand{\XRSolGain}{5.9}
\newcommand{\XRSolDiD}{5.5}\newcommand{\XRSolDiDLo}{2.3}\newcommand{\XRSolDiDHi}{10.0}
\newcommand{\XRAstraCtlGain}{2.1}
\newcommand{\XRAstraGain}{4.9}
\newcommand{\XRAstraDiD}{2.8}\newcommand{\XRAstraDiDLo}{$-$0.5}\newcommand{\XRAstraDiDHi}{8.1}
\newcommand{\XRSolDiDNoOutlier}{3.7}\newcommand{\XRAstraDiDNoOutlier}{0.5}
\newcommand{\XRAstraFinalDiff}{0.2}
\newcommand{\XRSolTouchedShown}{68}\newcommand{\XRSolTouchedCtl}{14}
\newcommand{\XRAstraTouchedShown}{57}\newcommand{\XRAstraTouchedCtl}{19}

\input{fig/data/redo/models_qual_macros}  
\input{fig/data/supp/why_fallback_macros}  

\input{fig/data/supp/weighting_macros}                 

\input{fig/data/supp/v23_rescore_macros}  
\input{fig/data/controls_v23/known_errors_text_macros}  

\makeatletter
\newcommand{\NoPlus}[1]{\begingroup\protected@edef\np@x{#1}\expandafter\np@strip\np@x\relax\endgroup}
\def\np@strip#1#2\relax{\if+#1\else#1\fi#2}
\def\eff@first#1#2\relax{\def\eff@c{#1}}
\newcommand{\EffCell}[1]{\begingroup
  \protected@edef\eff@m{\csname #1Mean\endcsname}%
  \protected@edef\eff@lo{\csname #1Lo\endcsname}\protected@edef\eff@hi{\csname #1Hi\endcsname}%
  \expandafter\eff@first\eff@lo\relax\let\eff@a\eff@c
  \expandafter\eff@first\eff@hi\relax\let\eff@b\eff@c
  \ifx\eff@a\eff@b\expandafter\eff@fmt\eff@m\relax
  \else\expandafter\eff@fmt\eff@m\relax\fi
  \endgroup}
\def\eff@fmt#1#2\relax{\if+#1\else\if-#1$-$\else#1\fi\fi#2}
\makeatother
\newcommand{\RoundTwo}[1]{\fpeval{round(#1,2)}}

\newif\ifPaperBreadthReady   \PaperBreadthReadyfalse    

\providecommand{\WhyPoseOffsetDeg}{\Prov{1}}

\providecommand{\PaperTerraVideoSimG}{\Prov{48.5}}
\providecommand{\PaperTerraVideoImageSimG}{\Prov{45.3}}
\providecommand{\PaperTerraEstimatedSimG}{\Prov{49.8}}
\providecommand{\PaperTerraPointsSimG}{\Prov{73.4}}
\providecommand{\PaperTerraOracleSimG}{\Prov{74.9}}
\providecommand{\PaperTerraMeasuredSimG}{\Prov{64.8}}
\providecommand{\PaperTerraImageSimG}{\Prov{64.7}}
\providecommand{\PaperSolVideoImageG}{\Prov{50.8}}
\providecommand{\PaperSolVideoSimG}{\Prov{48.2}}
\providecommand{\PaperSolVideoImageSimG}{\Prov{53.9}}
\providecommand{\PaperSolEstimatedSimG}{\Prov{54.5}}
\providecommand{\PaperSolPointsSimG}{\Prov{78.9}}
\providecommand{\PaperSolOracleSimG}{\Prov{82.7}}
\providecommand{\PaperSolMeasuredSimG}{\Prov{66.5}}
\providecommand{\PaperSolImageSimG}{\Prov{66.0}}
\providecommand{\PaperAstraVideoImageG}{\Prov{52.7}}
\providecommand{\PaperAstraOracleG}{\Prov{86.3}}
\providecommand{\PaperAstraVideoSimG}{\Prov{54.1}}
\providecommand{\PaperAstraVideoImageSimG}{\Prov{55.2}}
\providecommand{\PaperAstraEstimatedSimG}{\Prov{71.2}}
\providecommand{\PaperAstraPointsSimG}{\Prov{84.5}}
\providecommand{\PaperAstraOracleSimG}{\Prov{87.6}}
\providecommand{\PaperAstraMeasuredSimG}{\Prov{80.3}}
\providecommand{\PaperAstraImageSimG}{\Prov{78.6}}

\input{fig/entry_layout_switch}

%% file: fig/data/main_v3/paper_macros.tex
\input{fig/data/frozen_v2/final_macros}

\input{fig/data/diagnostics_v23/diag_macros}

\input{fig/data/derived_v23/derived_ranges}

\input{fig/data/read_audit_counts_20261113/macros}
\input{fig/data/body_numbers_20261113/macros}

\ExplSyntaxOn
\newcommand{\PaperFrozenPercent}[1]{\fp_eval:n{round(100*(#1),2)}}
\ExplSyntaxOff
\newcommand{\PaperSceneN}{30}

\newcommand{\PaperTerraVideoScaleError}{\DGTerraVideoScaleError}

\newcommand{\PaperTerraPointsScaleError}{\DGTerraPointsScaleError}

\newcommand{\PaperSolVideoScaleError}{\DGSolVideoScaleError}

\newcommand{\PaperAstraVideoScaleError}{\DGAstraVideoScaleError}

\newcommand{\PaperAstraPointsScaleError}{\DGAstraPointsScaleError}

\newcommand{\PaperTerraScaleGapMean}{\DGScaleGapTerraMean}

\newcommand{\PaperAstraScaleGapMean}{\DGScaleGapAstraMean}

\newcommand{\PaperCommonN}{\DGPoissonGapTerraRigidN}
\newcommand{\PaperCommonPoissonG}{\FTMeanTerraPoissonRigidG}

\newcommand{\PaperTerraCommonBudget}{3}

\newcommand{\PaperTerraPoissonGapMean}{\DGPoissonGapTerraRigidMean}

\newcommand{\PaperSolPoissonGapMean}{\DGPoissonGapSolRigidMean}

\newcommand{\PaperAstraCommonBudget}{39}

\newcommand{\PaperAstraPoissonGapMean}{\DGPoissonGapAstraRigidMean}

\newcommand{\PaperAstraEstimatedG}{\FTMeanAstraCThreeRigidG}

\newcommand{\PaperAstraEstimatedSimG}{\FTMeanAstraCThreeSimG}

\newcommand{\PaperAstraImageG}{\FTMeanAstraCFourRigidG}

\newcommand{\PaperAstraImageSimG}{\FTMeanAstraCFourSimG}

\newcommand{\PaperAstraMeasuredG}{\FTMeanAstraCOneRigidG}

\newcommand{\PaperAstraMeasuredSimG}{\FTMeanAstraCOneSimG}

\newcommand{\PaperAstraOracleDeltaMean}{\FTETwoARigidGMean}

\newcommand{\PaperAstraOracleG}{\FTMeanAstraCTwoORigidG}

\newcommand{\PaperAstraOracleSimG}{\FTMeanAstraCTwoOSimG}

\newcommand{\PaperAstraPointsDeltaMean}{\FTATwoRigidGMean}

\newcommand{\PaperAstraPointsG}{\FTMeanAstraCTwoRigidG}

\newcommand{\PaperAstraPointsSimG}{\FTMeanAstraCTwoSimG}

\newcommand{\PaperAstraScaleSlope}{\FTEThreeASlopeMean}

\newcommand{\PaperAstraVideoG}{\FTMeanAstraCZeroRigidG}

\newcommand{\PaperAstraVideoImageG}{\FTMeanAstraCZeroVRigidG}

\newcommand{\PaperAstraVideoImageSimG}{\FTMeanAstraCZeroVSimG}

\newcommand{\PaperAstraVideoSimG}{\FTMeanAstraCZeroSimG}

\newcommand{\PaperLunaPointsG}{\FTMeanLunaCTwoRigidG}

\newcommand{\PaperLunaPointsSimG}{\FTMeanLunaCTwoSimG}

\newcommand{\PaperLunaVideoG}{\FTMeanLunaCZeroRigidG}

\newcommand{\PaperLunaVideoImageG}{\FTMeanLunaCZeroVRigidG}

\newcommand{\PaperLunaVideoImageSimG}{\FTMeanLunaCZeroVSimG}

\newcommand{\PaperLunaVideoSimG}{\FTMeanLunaCZeroSimG}
\newcommand{\PaperPoissonThirtyG}{\FTMeanTerraPoissonRigidG}
\newcommand{\PaperPoissonThirtySimG}{\FTMeanTerraPoissonSimG}

\newcommand{\PaperPooledVideoImageDeltaMean}{\FTPoolRigidRigidGMean}

\newcommand{\PaperSolEstimatedG}{\FTMeanSolCThreeRigidG}

\newcommand{\PaperSolEstimatedSimG}{\FTMeanSolCThreeSimG}

\newcommand{\PaperSolImageG}{\FTMeanSolCFourRigidG}

\newcommand{\PaperSolImageSimG}{\FTMeanSolCFourSimG}

\newcommand{\PaperSolMeasuredG}{\FTMeanSolCOneRigidG}

\newcommand{\PaperSolMeasuredSimG}{\FTMeanSolCOneSimG}

\newcommand{\PaperSolOracleG}{\FTMeanSolCTwoORigidG}

\newcommand{\PaperSolOracleSimDeltaMean}{\FTETwoSSimGMean}

\newcommand{\PaperSolOracleSimG}{\FTMeanSolCTwoOSimG}

\newcommand{\PaperSolPointsDeltaMean}{\FTSOneRigidGMean}

\newcommand{\PaperSolPointsG}{\FTMeanSolCTwoRigidG}

\newcommand{\PaperSolPointsSimG}{\FTMeanSolCTwoSimG}

\newcommand{\PaperSolScaleSlope}{\FTEThreeSSlopeMean}

\newcommand{\PaperSolVideoG}{\FTMeanSolCZeroRigidG}

\newcommand{\PaperSolVideoImageG}{\FTMeanSolCZeroVRigidG}

\newcommand{\PaperSolVideoImageSimG}{\FTMeanSolCZeroVSimG}

\newcommand{\PaperSolVideoSimG}{\FTMeanSolCZeroSimG}

\newcommand{\PaperTerraEstimatedG}{\FTMeanTerraCThreeRigidG}

\newcommand{\PaperTerraEstimatedSimG}{\FTMeanTerraCThreeSimG}

\newcommand{\PaperTerraImageG}{\FTMeanTerraCFourRigidG}

\newcommand{\PaperTerraImageSimG}{\FTMeanTerraCFourSimG}

\newcommand{\PaperTerraMeasuredG}{\FTMeanTerraCOneRigidG}

\newcommand{\PaperTerraMeasuredSimG}{\FTMeanTerraCOneSimG}

\newcommand{\PaperTerraOracleDeltaMean}{\FTETwoTRigidGMean}

\newcommand{\PaperTerraOracleG}{\FTMeanTerraCTwoORigidG}

\newcommand{\PaperTerraOracleSimDeltaMean}{\FTETwoTSimGMean}

\newcommand{\PaperTerraOracleSimG}{\FTMeanTerraCTwoOSimG}

\newcommand{\PaperTerraPointsDeltaMean}{\FTTFourRigidGMean}

\newcommand{\PaperTerraPointsG}{\FTMeanTerraCTwoRigidG}

\newcommand{\PaperTerraPointsSimG}{\FTMeanTerraCTwoSimG}

\newcommand{\PaperTerraScaleSlopeN}{\FTEThreeTSlopeN}
\newcommand{\PaperTerraVideoG}{\FTMeanTerraCZeroRigidG}

\newcommand{\PaperTerraVideoImageG}{\FTMeanTerraCZeroVRigidG}

\newcommand{\PaperTerraVideoImageSimG}{\FTMeanTerraCZeroVSimG}

\newcommand{\PaperTerraVideoSimG}{\FTMeanTerraCZeroSimG}

\input{fig/data/supp_tables_20261113/macros}

%% file: fig/data/frozen_v2/final_macros.tex
\newcommand{\FTTOneRigidSMeanA}{50.9}

\newcommand{\FTTOneRigidVMeanA}{48.8}

\newcommand{\FTTOneRigidLMeanA}{30.6}

\newcommand{\FTTFourRigidGMean}{+18.0}

\newcommand{\FTSOneRigidGMean}{+18.2}

\newcommand{\FTSFiveRigidSMeanA}{56.0}

\newcommand{\FTSFiveRigidVMeanA}{50.9}

\newcommand{\FTSFiveRigidLMeanA}{36.8}

\newcommand{\FTETwoTRigidGMean}{+8.8}

\newcommand{\FTETwoTSimGMean}{+1.5}

\newcommand{\FTETwoSSimGMean}{+4.2}

\newcommand{\FTETwoARigidGMean}{+14.0}

\newcommand{\FTEThreeTSlopeN}{10}

\newcommand{\FTEThreeSSlopeMean}{+0.961}

\newcommand{\FTEThreeASlopeMean}{+0.927}

\newcommand{\FTATwoRigidGMean}{+6.9}

\newcommand{\FTASevenRigidSMeanA}{61.9}

\newcommand{\FTASevenRigidVMeanA}{56.8}

\newcommand{\FTASevenRigidLMeanA}{41.8}

\newcommand{\FTPoolRigidRigidGMean}{+2.6}

\newcommand{\FTMeanTerraCZeroRigidG}{41.1}
\newcommand{\FTMeanTerraCZeroSimG}{48.5}

\newcommand{\FTMeanTerraCZeroVRigidG}{43.3}
\newcommand{\FTMeanTerraCZeroVSimG}{45.3}

\newcommand{\FTMeanTerraCOneRigidG}{60.7}
\newcommand{\FTMeanTerraCOneSimG}{64.8}

\newcommand{\FTMeanTerraCTwoRigidG}{64.8}
\newcommand{\FTMeanTerraCTwoSimG}{73.4}

\newcommand{\FTMeanTerraCTwoORigidG}{73.6}
\newcommand{\FTMeanTerraCTwoOSimG}{74.9}

\newcommand{\FTMeanTerraCThreeRigidG}{46.8}
\newcommand{\FTMeanTerraCThreeSimG}{49.8}

\newcommand{\FTMeanTerraCFourRigidG}{61.5}
\newcommand{\FTMeanTerraCFourSimG}{64.7}

\newcommand{\FTMeanTerraPoissonRigidG}{75.9}
\newcommand{\FTMeanTerraPoissonSimG}{84.7}

\newcommand{\FTMeanSolCZeroRigidG}{46.2}
\newcommand{\FTMeanSolCZeroSimG}{48.2}

\newcommand{\FTMeanSolCZeroVRigidG}{50.8}
\newcommand{\FTMeanSolCZeroVSimG}{53.9}

\newcommand{\FTMeanSolCOneRigidG}{63.2}
\newcommand{\FTMeanSolCOneSimG}{66.5}

\newcommand{\FTMeanSolCTwoRigidG}{69.4}
\newcommand{\FTMeanSolCTwoSimG}{78.9}

\newcommand{\FTMeanSolCTwoORigidG}{81.3}
\newcommand{\FTMeanSolCTwoOSimG}{83.1}

\newcommand{\FTMeanSolCThreeRigidG}{51.3}
\newcommand{\FTMeanSolCThreeSimG}{54.5}

\newcommand{\FTMeanSolCFourRigidG}{63.9}
\newcommand{\FTMeanSolCFourSimG}{66.0}

\newcommand{\FTMeanAstraCZeroRigidG}{51.4}
\newcommand{\FTMeanAstraCZeroSimG}{53.4}

\newcommand{\FTMeanAstraCZeroVRigidG}{53.1}
\newcommand{\FTMeanAstraCZeroVSimG}{55.2}

\newcommand{\FTMeanAstraCOneRigidG}{75.9}
\newcommand{\FTMeanAstraCOneSimG}{78.8}

\newcommand{\FTMeanAstraCTwoRigidG}{73.1}
\newcommand{\FTMeanAstraCTwoSimG}{85.2}

\newcommand{\FTMeanAstraCTwoORigidG}{87.1}
\newcommand{\FTMeanAstraCTwoOSimG}{88.8}

\newcommand{\FTMeanAstraCThreeRigidG}{66.3}
\newcommand{\FTMeanAstraCThreeSimG}{71.9}

\newcommand{\FTMeanAstraCFourRigidG}{74.1}
\newcommand{\FTMeanAstraCFourSimG}{76.8}

\newcommand{\FTMeanLunaCZeroRigidG}{20.6}
\newcommand{\FTMeanLunaCZeroSimG}{30.6}

\newcommand{\FTMeanLunaCZeroVRigidG}{21.7}
\newcommand{\FTMeanLunaCZeroVSimG}{30.8}

\newcommand{\FTMeanLunaCTwoRigidG}{26.2}
\newcommand{\FTMeanLunaCTwoSimG}{33.4}

%% file: fig/data/diagnostics_v23/diag_macros.tex
\newcommand{\DGScaleGapTerraMean}{+8.6}

\newcommand{\DGScaleGapAstraMean}{+12.1}

\newcommand{\DGPoissonGapTerraRigidN}{30}
\newcommand{\DGPoissonGapTerraRigidMean}{+11.1}

\newcommand{\DGPoissonGapSolRigidMean}{+6.5}

\newcommand{\DGPoissonGapAstraRigidMean}{+2.8}

\newcommand{\DGPoissonGapAstraSimMean}{-0.6}

\newcommand{\DGTerraVideoScaleError}{10.4}

\newcommand{\DGTerraPointsScaleError}{2.9}

\newcommand{\DGSolVideoScaleError}{5.8}

\newcommand{\DGAstraVideoScaleError}{4.6}

\newcommand{\DGAstraPointsScaleError}{3.3}

%% file: fig/data/derived_v23/derived_ranges.tex
\newcommand{\DRMeasuredVsVideoMin}{17.0}
\newcommand{\DRMeasuredVsVideoMax}{24.5}

\newcommand{\DRScaleFitGainMin}{8.8}

%% file: fig/data/body_numbers_20261113/macros.tex
\newcommand{\BodyAppearanceSignificantN}{8}

%% file: fig/data/supp_tables_20261113/macros.tex
\input{tab/supp_l_sensitivity-macros}
\input{tab/supp_mirror-macros}
\input{tab/supp_mirror_effects-macros}
\input{tab/supp_appearance-macros}
\input{tab/supp_perfect_reader-macros}
\input{tab/supp_views-macros}
\input{tab/supp_shift-macros}
\input{tab/supp_shift_object-macros}
\input{tab/supp_height_slope-macros}
\input{fig/data/supp_tables_20261113/metadata_macros}

%% file: fig/data/main_v3/availability.tex
\newif\ifPaperPublicReady
\PaperPublicReadyfalse
\newif\ifPaperMeasurementValidated
\PaperMeasurementValidatedtrue  
\newif\ifPaperLunaRegisteredReady
\PaperLunaRegisteredReadytrue

%% file: tab/main_v3_table_macro_inputs.tex
\input{tab/main_v3_all_means-macros}

\input{tab/main_v3_components-macros}

\input{tab/main_v3_paired_effects-macros}
\input{tab/main_v3_direct_means-macros}
\input{tab/main_v3_resources-macros}

\input{tab/main_v3_dose-macros}
\input{tab/main_v3_common_gaps-macros}
\input{tab/working_v22_known_errors-macros}
\input{tab/main_v3_oracle_terra-macros}
\input{tab/main_v3_oracle_sol-macros}
\input{tab/main_v3_dose_per_scene-macros}

%% file: tab/main_v3_all_means-macros.tex
\newcommand{\AllMeansVThreeTerraCZeroGsim}{\FTMeanTerraCZeroSimG}

%% file: tab/main_v3_components-macros.tex
\newcommand{\ComponentsVThreeTerraCZeroS}{\FTTOneRigidSMeanA}
\newcommand{\ComponentsVThreeTerraCZeroV}{\FTTOneRigidVMeanA}
\newcommand{\ComponentsVThreeTerraCZeroL}{\FTTOneRigidLMeanA}

\newcommand{\ComponentsVThreeSolCZeroS}{\FTSFiveRigidSMeanA}
\newcommand{\ComponentsVThreeSolCZeroV}{\FTSFiveRigidVMeanA}
\newcommand{\ComponentsVThreeSolCZeroL}{\FTSFiveRigidLMeanA}

\newcommand{\ComponentsVThreeAstraCZeroS}{\FTASevenRigidSMeanA}
\newcommand{\ComponentsVThreeAstraCZeroV}{\FTASevenRigidVMeanA}
\newcommand{\ComponentsVThreeAstraCZeroL}{\FTASevenRigidLMeanA}

%% file: tab/main_v3_resources-macros.tex
\newcommand{\ResourceVThreeAstraCOneMeanTurns}{63.9}

\newcommand{\ResourceVThreeAstraCFourMeanTurns}{112.1}

%% file: fig/working_figure_assets.tex
\input{fig/data/redo/caption_macros}
\input{fig/data/redo/teaser_v10_macros}
\input{fig/data/redo/released_gpt_v23/macros}
\newif\ifWorkingTeaserPreview
\IfFileExists{fig/teaser_v13D.pdf}
  {}
  {\WorkingTeaserPreviewtrue}

\input{fig/data/redo/fig02_r2_c_macros}

\newif\ifWorkingProtocolPreview
\IfFileExists{fig/fig02_r4_45663154.pdf}
  {\newcommand{\WorkingProtocolFigure}{fig/fig02_r4_45663154}}
  {\newcommand{\WorkingProtocolFigure}{fig/fig_02_pipeline}\WorkingProtocolPreviewtrue}

\newif\ifWorkingReleasedPreview
\IfFileExists{fig/fig_03_ls_within_v2.pdf}
  {\WorkingReleasedPreviewtrue}
  {\WorkingReleasedPreviewtrue}

\newif\ifWorkingEvidencePreview
\IfFileExists{fig/fig45_evidence_v9_e88.pdf}
  {\newcommand{\WorkingEvidenceFigure}{fig/fig45_evidence_v9_e88}}
  {\newcommand{\WorkingEvidenceFigure}{fig/preview_fig_05_evidence_v3}\WorkingEvidencePreviewtrue}

\newif\ifWorkingModelsPreview
\IfFileExists{fig/fig_06_models_v3.pdf}
  {}
  {\WorkingModelsPreviewtrue}

\newif\ifWorkingDiagnosticsPreview
\WorkingDiagnosticsPreviewtrue
\IfFileExists{fig/fig05_why_v13.pdf}
  {%
   \ifPaperMeasurementValidated\WorkingDiagnosticsPreviewfalse\fi}
  {}

\newif\ifWorkingMatrixPreview
\IfFileExists{fig/fig_scene_matrix_full_v3.pdf}
  {}
  {\WorkingMatrixPreviewtrue}

\IfFileExists{fig/loop_v12.pdf}
  {\newcommand{\WorkingLoopFigure}{fig/loop_v12}}{}

%% file: fig/data/redo/fig02_r2_c_macros.tex
\newcommand{\FigTwoG}{38.7}

%% file: fig/data/section7/section7_report_test_macros.tex
\newcommand{\WRRigDiDMean}{4.3}

\newcommand{\WRRigGapCtl}{12.1}

\newcommand{\WRTouchedShownPct}{39}

\newcommand{\WRTouchedCtlPct}{10}

%% file: fig/data/section7/section7_intervention_macros.tex
\newcommand{\WIRepTwoDiDMean}{3.9}
\newcommand{\WIRepTwoDiDLo}{1.5}
\newcommand{\WIRepTwoDiDHi}{6.7}

\newcommand{\WIRepTwoGainMean}{5.2}

\newcommand{\WIRepTwoCtlGainMean}{1.2}

%% file: fig/data/supp/why_fallback_macros.tex
\newcommand{\WhyPoseOffsetDeg}{1}  

%% file: fig/data/supp/weighting_macros.tex
\newcommand{\WSShellG}{36.3}

%% file: fig/data/supp/v23_rescore_macros.tex
\newcommand{\VRRelAcrossRho}{0.79}

\newcommand{\VRRelWithinMedian}{0.10}

\newcommand{\VRRelConfigsMain}{54}

\newcommand{\VRRelTopG}{50.7}

\newcommand{\VRRelSecondG}{49.5}

%% file: fig/data/controls_v23/known_errors_text_macros.tex
\newcommand{\KEScaleDG}{58.5}

%% file: authors.tex
\author[1]{Ziyang Long}
\author[2]{Yen-Liang Lin}
\author[3]{Shifan Zhu}
\author[4]{Allen Tu}
\author[2]{Meher Gitika Karumuri}
\author[5]{Chenqi Li}
\author[6]{Srikumar Sastry}
\author[7]{Sarosij Bose}
\author[8]{Mellon M. Zhang}
\author[9,1]{Hsin-Jung Yang}
\affiliation[1]{University of California, Los Angeles}
\affiliation[2]{Amazon}
\affiliation[3]{Department of Computer Science, University of Massachusetts Amherst}
\affiliation[4]{Department of Computer Science, University of Maryland}
\affiliation[5]{University of Oxford}
\affiliation[6]{Washington University in St. Louis}
\affiliation[7]{University of California, Riverside}
\affiliation[8]{Georgia Institute of Technology}
\affiliation[9]{Cedars-Sinai Medical Center}

%% file: sec/0_abstract.tex
Vision-language agents can reconstruct rooms from video as editable 3D scenes, but convincing renders do not establish geometric accuracy.
We introduce a scan-referenced evaluation and use controlled experiments to study reconstruction accuracy, the role of geometric inputs, and agents' ability to verify and correct their scenes.
Our geometric score measures surface, depth, and furniture placement agreement with laser scans of \PaperSceneN{} real rooms on a 0--100 scale.
The strongest tested agent, GPT-6 Astra, scores \PaperAstraVideoG{} from video alone, while a dense reconstruction from the same video scores \PaperPoissonThirtyG{}.
Given the point cloud used for that reconstruction, Astra produces an editable scene that scores within three points of it.
Agents also inherit input errors: rescaling the supplied geometry changes scene scale almost proportionally.
During revision, viewing renders improves appearance more consistently than geometry.
Logs of their tool calls show that agents repeatedly measure their inputs but rarely compare them with the scenes they build.
To provide this missing check, we introduce a geometric error report that compares each saved scene with the agent's input points and lists the largest discrepancies.
Without new geometric evidence, the report raises the revision gains of GPT-6.1 Terra and Sol by \WIRepTwoDiDMean{} and \XRSolDiD{} points over concurrent controls.

%% file: fig/main_v3_teaser.tex
\begin{figure}[t]
\centering
\includegraphics[width=\linewidth]{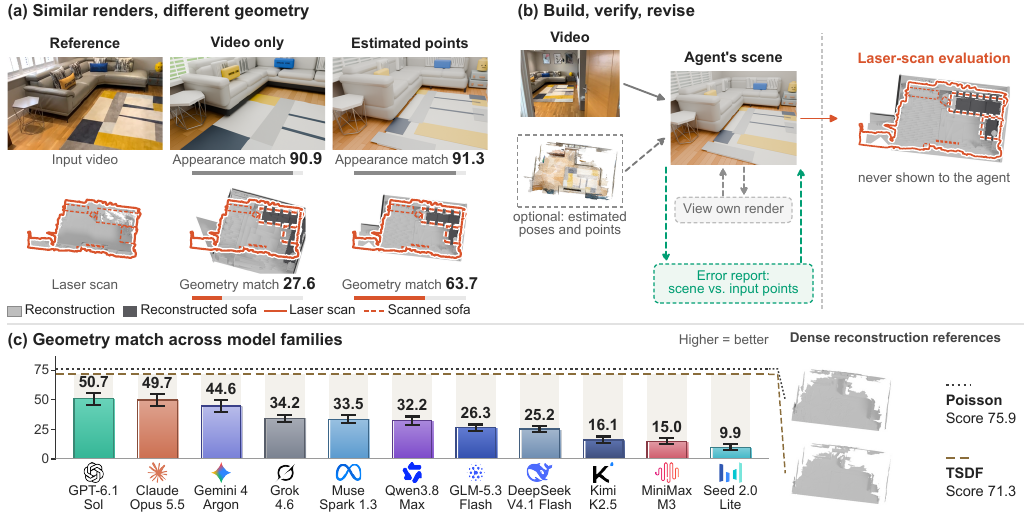}
\input{fig/entry_overview_caption}
\caption{\EntryOverviewCaption}
\label{fig:interventions-main}
\end{figure}

%% file: fig/entry_overview_caption.tex
\newcommand{\EntryOverviewCaption}{%
\textbf{Geometric evaluation of agent-built scenes.}
(a) Two GPT-6.1 Terra reconstructions of one room. \emph{Appearance match} compares renders with the video; \emph{Geometry match} compares geometry with the laser scan (both 0--100). Top views: registered geometry; outlines: scan and scanned sofa (dashed).
(b) Construction from video and optional estimated geometry; dashed: optional checks, including the error report of Section~\ref{sec:report-test}. Scan used only for evaluation.
(c) Publicly released reconstructions~\cite{tang2026bvb}: leading configuration per family; room means with 95\% intervals. Lines: dense TSDF~\cite{curless1996volumetric} and Poisson~\cite{kazhdan2013screened} references from video-estimated points, with example-room surfaces.}

%% file: sec/1_intro.tex
\section{Introduction}
\label{sec:intro}
Reconstructing rooms in 3D for furniture placement or robot navigation requires accurate geometry, not just convincing renders.
Suppose we ask an agent to rebuild a living room from a video.
The sofa, rug, and window may look right in its renders, but a cabinet that fits between the reconstructed sofa and wall may not fit in the real room.

Vision-language agents build these scenes by writing Blender programs
that define named, editable
objects~\cite{tang2026bvb,fang2026blenderworld,shen20264dcodebench}.
They can render, inspect, and revise their scenes.
However, evaluation focuses largely on the renders: the
BVB benchmark, for example, scores reconstructions by question-answer
retention and similarity between rendered and reference
videos~\cite{tang2026bvb}.
Looking right is not being right.
In Figure~\ref{fig:interventions-main}a, two reconstructions of one
room have nearly the same render score, while their agreement with
the room's laser scan differs by more than a factor of two.

A render score also cannot tell whether an agent misjudges geometry
from the video, loses supplied geometry when creating editable
objects, or fails to check the result.
These failures need different remedies, so we study each.

We therefore measure geometry directly and vary the agent's
geometric evidence and feedback one at a time, keeping the harness,
prompt, and budget fixed.
Replays of saved scenes and logs of tool calls show what agents
revise and verify.
Our contributions answer three questions, each raised by the answer
to the one before:
\begin{enumerate}[wide=0pt,label=\textbf{Q\arabic*.}]
\item[\textbf{Q1.}] \textbf{How accurately do agents reconstruct rooms from video alone?}
We introduce a geometric score that compares surfaces, per-view
depths, and furniture placement with laser scans of \PaperSceneN{}
real rooms from ARKitScenes~\cite{baruch2021arkitscenes}.
Controlled edits show that it detects errors render similarity misses.
On its 0--100 scale, the strongest tested agent scores
\PaperAstraVideoG{} versus \PaperPoissonThirtyG{} for a dense
reconstruction of that video.

\item[\textbf{Q2.}] \textbf{Where does geometric accuracy come from?}
Our controlled study shows that accuracy depends largely on supplied geometry.
Given estimated poses and points, the strongest agent comes within
three points of a dense surface built from the same points while
producing an editable scene.
Agents also inherit the supplied scale, even when it is wrong.

\item[\textbf{Q3.}] \textbf{Can agents verify and correct what they build?}
Viewing renders improves appearance more consistently
than geometry, and logs of their tool calls show that agents measure
their inputs but rarely compare them with their scenes.
We introduce a geometric error report that compares scenes with the
agent's own points.
It raises the revision gains of GPT-6.1 Terra and Sol by
\WIRepTwoDiDMean{} and \XRSolDiD{} points over concurrent controls.
\end{enumerate}

Together, these results show that building a scene and verifying it
are separate abilities.
The strongest tested agent converts supplied geometry into an editable
scene with little loss, so its accuracy is bounded by the errors of
that geometry.
Other agents revise without checking their scenes against that
geometry.
Our error report makes the check for them and improves their
accuracy.

%% file: sec/2_related.tex
\begin{figure*}[t]
  \centering
  \includegraphics{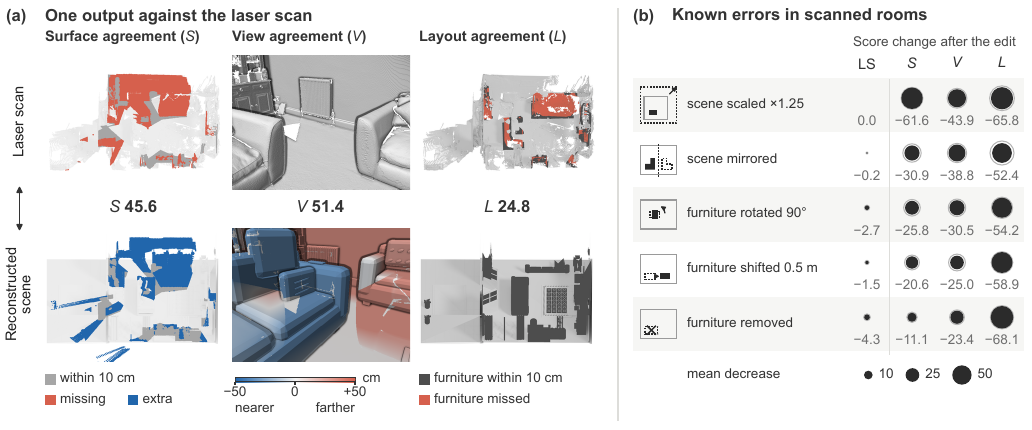}
  \caption{\textbf{Geometric scoring and known-error controls.}
  (a) A video-only Terra scene after rigid registration ($G=\FigTwoG{}$). $S$: missing scan surfaces (red) and excess scene surfaces (blue). $V$: signed scene-minus-scan depth. $L$: scanned furniture with (dark) or without (red) nearby scene surfaces. Scores aggregate the capture; views are illustrative.
  (b) Changes in video--render latent similarity (LS$\times100$) and geometric components after known edits to three scan-derived scenes. Disc area: mean decrease; ring: largest decrease.}
  \label{fig:metrics}
\end{figure*}

\section{Related Work}
\label{sec:related}

\paragraph{Agentic scene reconstruction.}
General-purpose agents that write scene programs are often evaluated
using render similarity, visual judgments or task-specific
checks~\cite{tang2026bvb,fang2026blenderworld,shen20264dcodebench}.
Systems that align CAD models to scans or produce structured scene
descriptions are evaluated against 3D annotations~\cite{avetisyan2019scan2cad,avetisyan2024scenescript}.
LEGO-Bench evaluates visible-surface geometry in single-image
reconstructions~\cite{li2026lego}, while SceneActBench evaluates multi-view furniture reconstruction and other 3D actions against hidden geometric ground truth~\cite{zhao2026sceneactbench}.
Our study evaluates visible geometry, furniture placement and metric
scale against room-scale laser scans, with controlled changes to
geometric inputs and feedback.

\paragraph{Feedback and verification.}
Visual feedback supports iterative refinement of scene
programs~\cite{hu2024scenecraft,huang2024blenderalchemy}.
LEGO-Anything documents regressive edits and unreliable
self-evaluation, and uses image-derived residuals and rollback to
refine single-image reconstructions~\cite{li2026lego}.
Programmatic evaluation also covers scene-state repair and physical
validity~\cite{ye2026code4scene,luo2026scenerig}.
We isolate image viewing in matched runs and test feedback that reports
discrepancies between the scene and the agent's existing points.

\paragraph{Geometric evaluation.}
Laser-referenced precision, recall and F-score are standard measures
for multi-view reconstruction~\cite{knapitsch2017tanks, yeshwanth2023scannet}.
Feed-forward methods estimate cameras and points from images or
video~\cite{wang2024dust3r,wang2025vggt,lin2026depth}, providing the
geometric evidence our agents receive.
We adapt these measurements to executable scene programs,
whose frame and scale are not fixed, adding an object-layout term
and registrations that distinguish scale and placement errors.

%% file: sec/3_protocol.tex
\section{Evaluation and Study Design}
\label{sec:protocol}

Agents reconstruct rooms from walkthrough videos by writing Blender
programs.
We export the scene geometry, register it to a laser scan of the same room,
and then measure surface agreement $S$, per-view depth agreement $V$,
and furniture placement $L$.
Figure~\ref{fig:metrics} illustrates the components and their sensitivity
to known errors.
Our experiments vary the geometric evidence and feedback agents receive
to evaluate their effects on reconstruction accuracy.

\subsection{Geometric Scoring}
All three components are computed over the region where the laser scan
has valid depth inside the room.
Missing scene surfaces count as failures.
For surface agreement $S$, we voxelize the reconstructed surfaces and reference scan at
2\,cm resolution, then average F-scores over 5, 10, and 20\,cm thresholds
to penalize missing or excess geometry.
View agreement $V$ combines per-view depth accuracy at 10\,cm with a depth-boundary
F-score that excludes specular regions detected by SAM3
\cite{carion2026sam} to avoid artifacts from reflections.
Layout agreement $L$ compares the scanned surfaces inside annotated furniture boxes
with the scene's non-structural surfaces using an F-score at 10\,cm.
Scoring furniture separately prevents the larger surfaces of walls and
floors from overshadowing placement errors.

Each component is normalized by its corresponding real-scan score to
give each room a reference level of one.
We define the overall geometric score as $G=100\sqrt[3]{SVL}$, capping
each component at one to keep $G$ on a 0--100 scale.
The geometric mean gives a high score only when all three components
show strong agreement.
To compare visual similarity with geometric accuracy, we also report
BVB's latent similarity (LS) as a 0--1 measure of agreement between
reference and rendered videos in learned feature space
\cite{tang2026bvb}.

\paragraph{Registration.}
The scene and scan may use different coordinate frames.
We fit a single gravity-aligned rigid transform to align the scene's
visible surfaces with the scan, preserving relative placement, metric
scale, and handedness.
Unless stated otherwise, $G$ denotes the score after this rigid fit.
Allowing a uniform scale gives $G_{\mathrm{s}}$.
Both fits minimize a truncated symmetric Chamfer distance rather than
$G$, so $G_{\mathrm{s}}-G$ is a scale-error diagnostic, not an exact
decomposition.
When camera poses are supplied, $G_{\mathrm{direct}}$ scores the scene
in their coordinate frame without geometric fitting.

\paragraph{Known-error controls.}
\label{sec:validation}
We apply controlled geometric edits to three scan-derived scenes to
compare the sensitivity of $G$ and render similarity.
Scaling a scene and its camera together by 1.25 leaves the renders,
and therefore their similarity score, unchanged, yet lowers $G$ by
\KEScaleDG{} points.
Figure~\ref{fig:metrics}b shows that mirroring the scene with its
camera, or rotating, shifting, or removing furniture, also lowers $G$
sharply while render similarity barely moves.
The components respond to different errors: removing furniture, for
example, lowers $L$ far more than $S$.
\subsection{Study Design}
\label{sec:study-design}
Our study uses \PaperSceneN{} rooms from ARKitScenes with RGB
walkthrough videos, laser depth scans, camera trajectories, and annotated 3D boxes
\cite{baruch2021arkitscenes,yang2025thinking}.
Agents run in our own harness, a minimal tool loop: a Blender sandbox
with a shell and frame extraction, plus image inspection in viewing
conditions.
They build scenes from primitives.
We evaluate GPT-6 Luna and Astra and GPT-6.1 Terra and
Sol~\cite{openai2026models}.
Terra, Sol, and Astra completed every condition and are our main models.
Terra and Sol receive \$\PaperTerraCommonBudget{} per scene and Astra
\$\PaperAstraCommonBudget{}, which at Astra's prices buys the same
number of tokens as Terra's \$\PaperTerraCommonBudget{}.
The budgets do not bind, as every run ends when the agent declares
the scene done.
Most runs leave over half their budget unspent.

We run each model once per room and condition, keeping its harness,
prompt, and budget fixed; unless noted, results cover all
\PaperSceneN{} rooms.
The conditions in Table~\ref{tab:conditions} vary the geometric
evidence supplied and whether agents can inspect their own renders.
DA3 \cite{lin2026depth} provides estimated camera poses and 3D points,
while oracle points are derived from ground-truth laser depth.
Measured camera poses come from ARKit tracking and typically agree
with video frames to about \WhyPoseOffsetDeg{}$^\circ$.
A separate experiment in Section~\ref{sec:report-test} tests feedback on
discrepancies between the scene and its supplied points.
We estimate intervention effects from paired comparisons within rooms
and obtain 95\% confidence intervals by bootstrapping rooms.

%% file: sec/5_results.tex
\section{How Accurate Are the Reconstructions?}
\label{sec:results}

From video alone, Terra, Sol, and Astra reach $G$ of
\PaperTerraVideoG{} to \PaperAstraVideoG{}.
For reference, a dense Poisson surface~\cite{kazhdan2013screened} built from points estimated
from the same video scores \PaperPoissonThirtyG{}; the laser scan's
walls, floor, and ceiling alone score \WSShellG{}.
Table~\ref{tab:video-components} shows that furniture placement is the
weakest component for all three models.

Scale errors explain part of this shortfall.
Median absolute scale errors are \PaperTerraVideoScaleError{}\%,
\PaperSolVideoScaleError{}\%, and \PaperAstraVideoScaleError{}\% for
Terra, Sol, and Astra.
Fitting scale at evaluation raises Terra's $G$ from \PaperTerraVideoG{}
to \AllMeansVThreeTerraCZeroGsim{}, while Sol and Astra gain only about
two points each.
Thus, most of the shortfall remains after scale fitting.

\paragraph{Render similarity is a weak guide to individual scenes.}
We rescore more than 2{,}000 publicly available reconstructions
from BVB \cite{tang2026bvb} to compare render similarity with
geometric accuracy.
The best released Sol and Astra configurations achieve $G$ scores of
\VRRelTopG{} and \VRRelSecondG{}, comparable to our best video-only
result and well below the dense references from the same video.
Figure~\ref{fig:interventions-main}c shows leading released configurations
alongside the dense references.
Across \VRRelConfigsMain{} configurations, the Spearman correlation
between mean render similarity and mean $G$ is \VRRelAcrossRho{}.
Within configurations, however, the median correlation across rooms is
only \VRRelWithinMedian{}.
Render similarity therefore offers little guidance on a room's
geometric accuracy.

\input{tab/study_conditions_float}
\input{tab/video_components_float}

%% file: tab/study_conditions_float.tex
\begin{table}[t]
  \centering
  \caption{\textbf{Input conditions.} All conditions receive the same RGB video. Oracle input replaces points and poses together; rescaling transforms estimated points and poses jointly.}
  \label{tab:conditions}
  \small
  \setlength{\tabcolsep}{4pt}
  \begin{tabular}{@{}ll@{}}
    \toprule
    \textbf{Condition} & \textbf{Evidence or access} \\
    \midrule
    \multicolumn{2}{@{}l}{\emph{No supplied poses}} \\
    \quad Video only         & No additional input \\
    \quad Video + viewing    & May view its own renders \\
    \multicolumn{2}{@{}l}{\emph{Estimated from the video}} \\
    \quad Estimated camera   & Estimated camera poses \\
    \quad Estimated points   & Estimated poses and 3D points \\
    \quad Rescaled points    & Poses and points scaled by 0.75--1.25 \\
    \multicolumn{2}{@{}l}{\emph{Measured and oracle inputs}} \\
    \quad Oracle points      & Laser depth points and ARKit poses \\
    \quad Measured camera    & ARKit camera poses \\
    \quad Measured + viewing & ARKit poses and render viewing \\
    \bottomrule
  \end{tabular}
\end{table}

%% file: tab/video_components_float.tex
\begin{table}[t]
  \centering
  \caption{\textbf{Video-only geometric accuracy.} Mean scores after rigid registration, on a 0--100 scale. Poisson uses points estimated from the same video.}
  \label{tab:video-components}
  \small
  \setlength{\tabcolsep}{4pt}
  \begin{tabularx}{0.8\linewidth}{@{}l*{4}{>{\centering\arraybackslash}X}@{}}
    \toprule
    \textbf{Model} & $\boldsymbol{S}$ & $\boldsymbol{V}$ & $\boldsymbol{L}$ & $\boldsymbol{G}$ \\
    \midrule
    GPT-6.1 Terra & \ComponentsVThreeTerraCZeroS{} & \ComponentsVThreeTerraCZeroV{} & \ComponentsVThreeTerraCZeroL{} & \PaperTerraVideoG{} \\
    GPT-6.1 Sol   & \ComponentsVThreeSolCZeroS{} & \ComponentsVThreeSolCZeroV{} & \ComponentsVThreeSolCZeroL{} & \PaperSolVideoG{} \\
    GPT-6 Astra & \ComponentsVThreeAstraCZeroS{} & \ComponentsVThreeAstraCZeroV{} & \ComponentsVThreeAstraCZeroL{} & \PaperAstraVideoG{} \\
    \midrule
    \multicolumn{5}{@{}l}{\emph{Dense reference:} a Poisson surface scores $G$~\PaperPoissonThirtyG{}.} \\
    \bottomrule
  \end{tabularx}
\end{table}

%% file: sec/5b_controlled.tex
\section{What Determines Geometric Accuracy?}
\label{sec:controlled}

\begin{figure*}[t]
  \centering
  \includegraphics{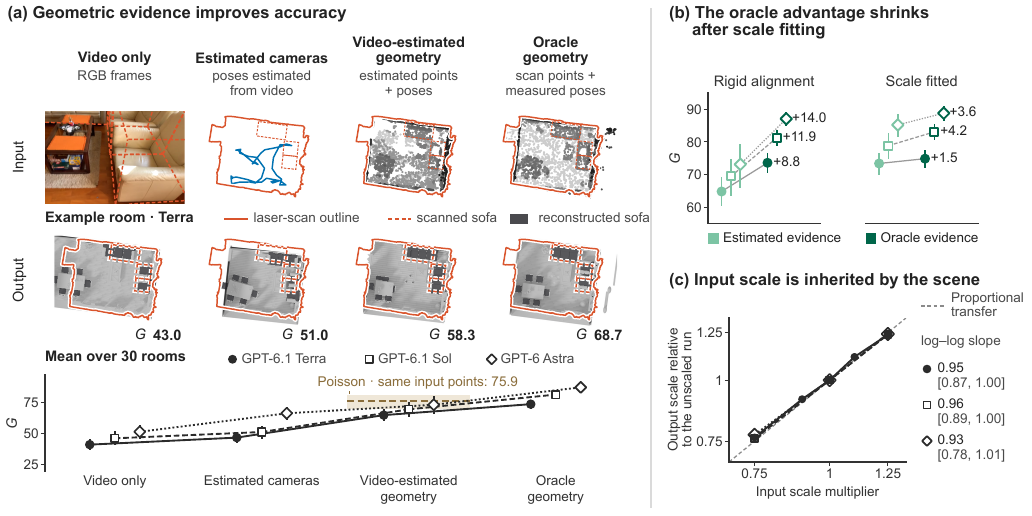}
  \caption{\textbf{Geometric inputs, conversion, and scale transfer.}
  (a) Inputs and Terra scenes for one room, with mean rigid $G$ below. All conditions receive video; oracle inputs replace estimated points and poses. Examples share a view, scale, and crop. Solid and dashed outlines mark the laser scan and scanned sofa; reconstructed sofas are dark. Poisson uses the same estimated points.
  (b) Paired oracle-minus-estimated gains before and after scale fitting.
  (c) Scene scale relative to the unscaled run as points and camera positions are jointly rescaled. Dashed line: proportional transfer; slopes: log--log fits. Intervals: 95\% room-bootstrap confidence intervals.}
  \label{fig:qualitative-main}
  \label{fig:evidence-measurements}
\end{figure*}

We next examine how supplied geometry improves accuracy,
how well agents preserve it in editable scenes, and whether they
inherit its scale errors.
Figure~\ref{fig:evidence-measurements} illustrates these comparisons,
and Tables~\ref{tab:answer} and~\ref{tab:paired} report accuracy across
conditions and mean paired changes within rooms.

\begin{table*}[t]
  \centering
  \caption{\textbf{Geometric accuracy by input and model.}
  Mean rigid $G$ and scale-fitted $G_{\mathrm{s}}$.}
  \label{tab:answer}
  \small
  \input{tab/main_v3_answer_display}
\end{table*}

\begin{table}[tb]
  \centering
  \caption{\textbf{Paired changes in geometric accuracy.}
  Entries are $\Delta G$, the mean within-room change in $G$ between
  the two conditions of each row.
  Scale fitting compares $G_{\mathrm{s}}$ with $G$ for scenes built
  from estimated points.}
  \label{tab:paired}
  \small
  \setlength{\tabcolsep}{5pt}
  \begin{tabular}{@{}lccc@{}}
    \toprule
    \textbf{Comparison} & \textbf{\shortstack{GPT-6.1\\Terra}} & \textbf{\shortstack{GPT-6.1\\Sol}} & \textbf{\shortstack{GPT-6\\Astra}} \\
    \midrule
    Measured camera vs.\ video   & \EffCell{FTTOneRigidG} & \EffCell{FTSFiveRigidG} & \EffCell{FTASevenRigidG} \\
    Estimated camera vs.\ video  & \EffCell{FTTTwoRigidG} & \EffCell{FTSSixRigidG} & \EffCell{FTAEightRigidG} \\
    Est.\ points vs.\ est.\ camera & \EffCell{FTTFourRigidG} & \EffCell{FTSOneRigidG} & \EffCell{FTATwoRigidG} \\
    Oracle vs.\ est.\ points     & \EffCell{FTETwoTRigidG} & \EffCell{FTETwoSRigidG} & \EffCell{FTETwoARigidG} \\
    \quad with scale fitted      & \EffCell{FTETwoTSimG} & \EffCell{FTETwoSSimG} & \EffCell{FTETwoASimG} \\
    Fitting scale, est.\ points  & \EffCell{DGScaleGapTerra} & \EffCell{DGScaleGapSol} & \EffCell{DGScaleGapAstra} \\
    \bottomrule
  \end{tabular}
\end{table}

\subsection{Metric Evidence Improves Accuracy}
\label{sec:evidence}

Table~\ref{tab:paired} shows that supplying measured camera poses
raises $G$ over video alone by
\DRMeasuredVsVideoMin{}--\DRMeasuredVsVideoMax{} points.
Estimated poses also raise mean $G$, and adding estimated points
increases it by a further
\NoPlus{\PaperTerraPointsDeltaMean} and
\NoPlus{\PaperSolPointsDeltaMean} points for Terra and Sol.
Astra reaches \PaperAstraEstimatedG{} with estimated poses alone
and gains another \NoPlus{\PaperAstraPointsDeltaMean} points with the point cloud.
The same agents build substantially more accurate scenes when given
poses and points, indicating that geometric inference from video is a
major source of reconstruction error.

\subsection{Preserving Geometry in Editable Scenes}
\label{sec:models}

We assess how well agents preserve supplied geometry by comparing
their editable scenes with an unstructured Poisson mesh built from the
same estimated points.
This mesh provides a reference for what the input points support, rather than
an upper bound or a competing editable reconstruction method.
Figure~\ref{fig:evidence-measurements}a shows its $G$ score of
\PaperCommonPoissonG{} over the same \PaperCommonN{} rooms.
Terra and Sol score \NoPlus{\PaperTerraPoissonGapMean} and
\NoPlus{\PaperSolPoissonGapMean} points below this reference.
Astra's deficit is
\NoPlus{\PaperAstraPoissonGapMean}
points. After fitting scale for both reconstructions, its mean
$G_{\mathrm{s}}$ is \fpeval{-\DGPoissonGapAstraSimMean} points higher
than the reference, but the 95\% confidence interval for this difference
includes zero.
Astra therefore approaches the dense reference while producing an
editable scene.
\ifPaperBreadthReady
\TODO{One sentence on the models from the other developers.}
\fi

\subsection{Scenes Inherit the Scale of Their Evidence}
\label{sec:input-error}

To measure how agents respond to scale errors, we jointly rescale supplied
points and camera positions by factors of 0.75--1.25 in
\PaperTerraScaleSlopeN{} rooms per model.
Image projections and the RGB video remain unchanged.
Figure~\ref{fig:evidence-measurements}c shows that scene scale tracks
supplied scale, with log--log slopes of
\RoundTwo{\PaperAstraScaleSlope}--\RoundTwo{\PaperSolScaleSlope}
across Terra, Sol, and Astra.
A slope of one indicates proportionality.
By comparison, their video-only scenes have median scale errors of
\PaperAstraVideoScaleError{}--\PaperTerraVideoScaleError{}\%, well
below the maximum 25\% perturbation.

Scenes built from unmodified estimated points also have median absolute
scale errors of
\PaperTerraPointsScaleError{}--\PaperAstraPointsScaleError{}\%.
Mean scale-fitting gains are
\NoPlus{\PaperTerraScaleGapMean}--\NoPlus{\PaperAstraScaleGapMean} points
in $G$ for the agents in Table~\ref{tab:paired} and
\DRScaleFitGainMin{} points for the Poisson reference.
For Astra, the scale-fitting gain exceeds the gap to Poisson, suggesting
that inherited scale error is a greater limitation than geometric loss
during scene construction.

Scale fitting also removes most of the gain from replacing estimated
points and poses with oracle input.
Figure~\ref{fig:evidence-measurements}b shows that laser depth points
and measured poses raise $G$ by
\NoPlus{\PaperTerraOracleDeltaMean}--\NoPlus{\PaperAstraOracleDeltaMean} points.
After scale fitting in both conditions, these gains shrink to
\NoPlus{\PaperTerraOracleSimDeltaMean}--\NoPlus{\PaperSolOracleSimDeltaMean} points.
Because oracle input changes points, poses, and coverage together,
we do not attribute the remaining gain to any single factor.
The scenes therefore gain accuracy from supplied geometry while
inheriting its scale errors.

%% file: tab/main_v3_answer_display.tex
\input{fig/data/table3_fill_v23/table3_fill_macros}
\newcommand{\TFcell}[1]{#1}
\setlength{\tabcolsep}{5pt}
\begin{tabular}{@{}l *{8}{w{c}{0.44in}}@{}}
\toprule
 & \multicolumn{2}{c}{\textbf{GPT-6 Luna}} & \multicolumn{2}{c}{\textbf{GPT-6.1 Terra}} & \multicolumn{2}{c}{\textbf{GPT-6.1 Sol}} & \multicolumn{2}{c}{\textbf{GPT-6 Astra}} \\
\cmidrule(lr){2-3}\cmidrule(lr){4-5}\cmidrule(lr){6-7}\cmidrule(l){8-9}
\textbf{Input condition} & $\boldsymbol{G}$ & $\boldsymbol{G_{\mathrm{s}}}$ & $\boldsymbol{G}$ & $\boldsymbol{G_{\mathrm{s}}}$ & $\boldsymbol{G}$ & $\boldsymbol{G_{\mathrm{s}}}$ & $\boldsymbol{G}$ & $\boldsymbol{G_{\mathrm{s}}}$ \\
\midrule
\multicolumn{9}{@{}l}{\emph{No supplied poses}} \\
\quad Video only & \PaperLunaVideoG{} & \PaperLunaVideoSimG{} & \PaperTerraVideoG{} & \PaperTerraVideoSimG{} & \PaperSolVideoG{} & \PaperSolVideoSimG{} & \PaperAstraVideoG{} & \PaperAstraVideoSimG{} \\
\quad Video + viewing & \PaperLunaVideoImageG{} & \PaperLunaVideoImageSimG{} & \PaperTerraVideoImageG{} & \PaperTerraVideoImageSimG{} & \PaperSolVideoImageG{} & \PaperSolVideoImageSimG{} & \PaperAstraVideoImageG{} & \PaperAstraVideoImageSimG{} \\
\addlinespace[2pt]
\multicolumn{9}{@{}l}{\emph{Estimated from the video}} \\
\quad Estimated camera & \TFcell{\TFMeanLunaEstimatedRigidG} & \TFcell{\TFMeanLunaEstimatedSimG} & \PaperTerraEstimatedG{} & \PaperTerraEstimatedSimG{} & \PaperSolEstimatedG{} & \PaperSolEstimatedSimG{} & \PaperAstraEstimatedG{} & \PaperAstraEstimatedSimG{} \\
\quad Estimated points & \PaperLunaPointsG{} & \PaperLunaPointsSimG{} & \PaperTerraPointsG{} & \PaperTerraPointsSimG{} & \PaperSolPointsG{} & \PaperSolPointsSimG{} & \PaperAstraPointsG{} & \PaperAstraPointsSimG{} \\
\addlinespace[2pt]
\multicolumn{9}{@{}l}{\emph{Measured and oracle inputs}} \\
\quad Oracle points & \TFcell{\TFMeanLunaOracleRigidG} & \TFcell{\TFMeanLunaOracleSimG} & \PaperTerraOracleG{} & \PaperTerraOracleSimG{} & \PaperSolOracleG{} & \PaperSolOracleSimG{} & \PaperAstraOracleG{} & \PaperAstraOracleSimG{} \\
\quad Measured camera & \TFcell{\TFMeanLunaMeasuredRigidG} & \TFcell{\TFMeanLunaMeasuredSimG} & \PaperTerraMeasuredG{} & \PaperTerraMeasuredSimG{} & \PaperSolMeasuredG{} & \PaperSolMeasuredSimG{} & \PaperAstraMeasuredG{} & \PaperAstraMeasuredSimG{} \\
\quad Measured + viewing & \TFcell{\TFMeanLunaImageRigidG} & \TFcell{\TFMeanLunaImageSimG} & \PaperTerraImageG{} & \PaperTerraImageSimG{} & \PaperSolImageG{} & \PaperSolImageSimG{} & \PaperAstraImageG{} & \PaperAstraImageSimG{} \\
\midrule
\multicolumn{9}{@{}l}{\emph{Dense reference:} a Poisson surface built from the same estimated points scores $G$~\PaperPoissonThirtyG{} and $G_{\mathrm{s}}$~\PaperPoissonThirtySimG{}.} \\
\bottomrule
\end{tabular}

%% file: fig/data/table3_fill_v23/table3_fill_macros.tex
\newcommand{\TFMeanLunaMeasuredN}{30}
\newcommand{\TFMeanLunaMeasuredRigidG}{26.7}
\newcommand{\TFMeanLunaMeasuredSimG}{33.6}
\newcommand{\TFMeanLunaMeasuredDirectG}{17.3}
\newcommand{\TFMeanLunaMeasuredMirHandG}{27.5}
\newcommand{\TFMeanLunaMeasuredMirGRuleG}{29.0}
\newcommand{\TFPzeroALunaMeasuredRuns}{30}
\newcommand{\TFPzeroALunaMeasuredRigidFail}{25}
\newcommand{\TFPzeroALunaMeasuredSimFail}{23}
\newcommand{\TFMeanLunaEstimatedN}{30}
\newcommand{\TFMeanLunaEstimatedRigidG}{24.3}
\newcommand{\TFMeanLunaEstimatedSimG}{31.9}
\newcommand{\TFMeanLunaEstimatedMirHandG}{25.2}
\newcommand{\TFMeanLunaEstimatedMirGRuleG}{27.2}
\newcommand{\TFPzeroALunaEstimatedRuns}{30}
\newcommand{\TFPzeroALunaEstimatedRigidFail}{23}
\newcommand{\TFPzeroALunaEstimatedSimFail}{24}
\newcommand{\TFMeanLunaImageN}{30}
\newcommand{\TFMeanLunaImageRigidG}{26.6}
\newcommand{\TFMeanLunaImageSimG}{33.7}
\newcommand{\TFMeanLunaImageDirectG}{17.4}
\newcommand{\TFMeanLunaImageMirHandG}{27.1}
\newcommand{\TFMeanLunaImageMirGRuleG}{28.2}
\newcommand{\TFPzeroALunaImageRuns}{30}
\newcommand{\TFPzeroALunaImageRigidFail}{26}
\newcommand{\TFPzeroALunaImageSimFail}{24}
\newcommand{\TFMeanLunaOracleN}{30}
\newcommand{\TFMeanLunaOracleRigidG}{29.1}
\newcommand{\TFMeanLunaOracleSimG}{34.0}
\newcommand{\TFMeanLunaOracleMirHandG}{29.5}
\newcommand{\TFMeanLunaOracleMirGRuleG}{31.4}
\newcommand{\TFPzeroALunaOracleRuns}{30}
\newcommand{\TFPzeroALunaOracleRigidFail}{18}
\newcommand{\TFPzeroALunaOracleSimFail}{17}
\newcommand{\TFMeanGrokMeasuredN}{10}
\newcommand{\TFMeanGrokMeasuredRigidG}{41.9}
\newcommand{\TFMeanGrokMeasuredSimG}{47.2}
\newcommand{\TFMeanGrokMeasuredDirectG}{29.9}
\newcommand{\TFMeanGrokMeasuredMirHandG}{41.9}
\newcommand{\TFMeanGrokMeasuredMirGRuleG}{43.5}
\newcommand{\TFPzeroAGrokMeasuredRuns}{10}
\newcommand{\TFPzeroAGrokMeasuredRigidFail}{4}
\newcommand{\TFPzeroAGrokMeasuredSimFail}{3}
\newcommand{\TFMeanGrokEstimatedN}{10}
\newcommand{\TFMeanGrokEstimatedRigidG}{38.3}
\newcommand{\TFMeanGrokEstimatedSimG}{46.4}
\newcommand{\TFMeanGrokEstimatedMirHandG}{38.3}
\newcommand{\TFMeanGrokEstimatedMirGRuleG}{40.2}
\newcommand{\TFPzeroAGrokEstimatedRuns}{10}
\newcommand{\TFPzeroAGrokEstimatedRigidFail}{4}
\newcommand{\TFPzeroAGrokEstimatedSimFail}{4}
\newcommand{\TFMeanGrokImageN}{10}
\newcommand{\TFMeanGrokImageRigidG}{43.2}
\newcommand{\TFMeanGrokImageSimG}{46.1}
\newcommand{\TFMeanGrokImageDirectG}{25.0}
\newcommand{\TFMeanGrokImageMirHandG}{43.2}
\newcommand{\TFMeanGrokImageMirGRuleG}{43.5}
\newcommand{\TFPzeroAGrokImageRuns}{10}
\newcommand{\TFPzeroAGrokImageRigidFail}{5}
\newcommand{\TFPzeroAGrokImageSimFail}{6}
\newcommand{\TFMeanGrokOracleN}{10}
\newcommand{\TFMeanGrokOracleRigidG}{64.0}
\newcommand{\TFMeanGrokOracleSimG}{62.4}
\newcommand{\TFMeanGrokOracleMirHandG}{64.0}
\newcommand{\TFMeanGrokOracleMirGRuleG}{64.0}
\newcommand{\TFPzeroAGrokOracleRuns}{10}
\newcommand{\TFPzeroAGrokOracleRigidFail}{0}
\newcommand{\TFPzeroAGrokOracleSimFail}{1}

%% file: sec/6_verification.tex
\section{Can Agents Verify and Correct Geometry?}
\label{sec:why}
\input{fig/data/section7/section7_macros}
\input{fig/data/section7/section7_trace_addendum_macros}  
\input{fig/data/section7/section7_firstfinal_macros}

Agents inherit errors in supplied geometry and can introduce further
errors during scene construction.
We compare runs with and without viewing renders to assess the effect
on geometry, while saved-scene replays and execution traces, the logs
of each agent's tool calls, show what agents revise and verify.
An error-report intervention tests whether feedback on scene-to-input
discrepancies improves geometry correction.

\input{fig/appearance_effects}

\begin{figure*}[t]
  \centering
  \ifdefined\WorkingLoopFigure
    \includegraphics{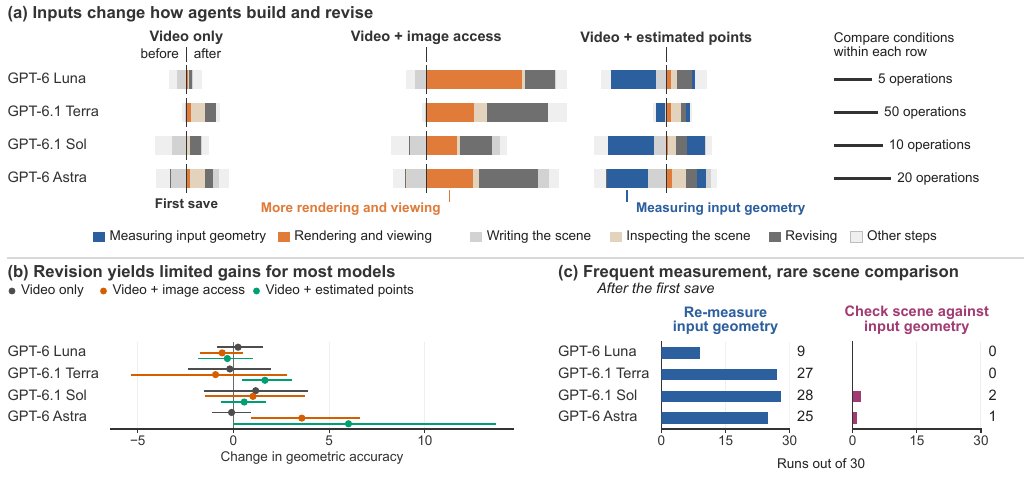}
  \else
    \fbox{\parbox{0.9\linewidth}{\centering\vspace{2.5em}
    Loop figure (pending)\vspace{2.5em}}}
  \fi
  \caption{\textbf{Construction, revision, and geometric checks.}
  (a) Mean operations before and after the first save, grouped by type rather than execution order. Scales differ by model. Estimated points include poses.
  (b) First-to-final change in rigid $G$ with 95\% room-bootstrap intervals, over multi-save runs with a scored first scene.
  (c) Estimated-point runs with input re-measurement or explicit scene-to-input comparison after the first save.}
  \label{fig:loop}
\end{figure*}

\subsection{Visual Checks Mostly Improve Appearance}
\label{sec:looking}

We assess visual verification by allowing agents to inspect their own renders.
Figure~\ref{fig:appearance-effects} contrasts the effects on appearance
and geometry, and Table~\ref{tab:viewing-effects} reports paired
changes in $G$.
\begin{table}[t]
  \centering
  \caption{\textbf{Effect of viewing renders on geometry.}
  Entries are $\Delta G$, the mean within-room change in $G$ when the
  agent can view its renders.
  $^\dagger$$G_{\mathrm{direct}}$ is the unregistered score.}
  \label{tab:viewing-effects}
  \small
  \setlength{\tabcolsep}{5pt}
  \begin{tabular}{@{}lccc@{}}
    \toprule
    \textbf{Camera input} & \textbf{\shortstack{GPT-6.1\\Terra}} & \textbf{\shortstack{GPT-6.1\\Sol}} & \textbf{\shortstack{GPT-6\\Astra}} \\
    \midrule
    None & \EffCell{FTTSevenRigidG} & \EffCell{FTSEightRigidG} & \EffCell{FTAOneZeroRigidG} \\
    Measured$^\dagger$ & \EffCell{FTTSixDirectG} & \EffCell{FTSFourDirectG} & \EffCell{FTAOneDirectG} \\
    \bottomrule
  \end{tabular}
\end{table}

\emph{With a measured camera}, viewing produces no detectable gain
in $G_{\mathrm{direct}}$ for any of the three models.
All three intervals include zero, and their upper bounds are small
compared with the gains from the camera itself.

\emph{Without a camera}, $G$ rises by
\NoPlus{\PaperPooledVideoImageDeltaMean} points on average across five
models, although only Sol's interval excludes zero.
The gain is small and partly depends on how left--right mirrored
reconstructions are scored.

Viewing affects appearance more consistently.
With a measured camera, all five appearance measures
move toward the video in all three models, and
\BodyAppearanceSignificantN{} of the 15 intervals exclude zero.
Astra's mean turn count rises from
\ResourceVThreeAstraCOneMeanTurns{} to
\ResourceVThreeAstraCFourMeanTurns{} per scene, without a corresponding improvement
in geometric accuracy.

\subsection{Revision Rarely Verifies Geometry}
\label{sec:inside-loop}

Execution traces show how agents revise their scenes and check geometry.
We label tool calls and final replies by the operations they perform: fetching frames, measuring input geometry, writing, rendering, inspecting, or revising a scene. Geometric checks explicitly compare the saved scene with input geometry; visual inspection alone does not count as a geometric check.
Figure~\ref{fig:loop} summarizes the reconstruction loop, changes in
accuracy during revision, and checks against the supplied geometry.

All models fetch frames, measure any supplied geometry, save a first
scene, and then render and revise it.
Figure~\ref{fig:loop}a shows that evidence lengthens initial construction.
Measuring estimated points accounts for about a third of model-call spending.
Astra also takes roughly three times as many turns to reach its first
save compared with video alone.
Viewing instead lengthens revision after the first save.

Unassisted revision adds little geometric accuracy for most models.
We replayed runs up to their first save and scored the recovered scenes.
Figure~\ref{fig:loop}b shows first-to-final changes of
\WFTerraVideoRigidMean{} to \WFSolVideoRigidMean{} points across
Terra, Sol, Astra, and Luna with video alone.
With estimated points, mean gains are \WFTerraPointsRigidMean{},
\WFSolPointsRigidMean{}, and \WFAstraPointsRigidMean{} points for
Terra, Sol, and Astra; only Astra gains substantially, and its gain
varies widely across rooms.
With viewing and no camera, Astra's mean revision gain is
\WFAstraVideoViewRigidMean{} points and its 95\% confidence interval
excludes zero.
Viewing prompts more changes to the set of objects, while
revisions that only move objects remain few.

Agents analyze supplied geometry but rarely use it to check their scenes.
Before the first save, they summarize points with histograms and quantiles.
Sol and Astra also fit planes, estimate local surface normals, and cluster points.
Placements traced to printed input measurements differ from those
measurements by a median of \WLTracedDiffMedianCm{}\,cm.
Figure~\ref{fig:loop}c shows that after the first save, each main model
re-measures inputs in most estimated-point runs but explicitly
compares the scene with them in at most two of 30 runs.
The broader audit finds explicit scene-to-input comparisons in only
\WLSceneVsPointsRuns{} of \WLPointRunsN{} point-cloud runs, based on
recorded operations rather than visual judgments.
Geometric analysis therefore guides construction but rarely serves as a check
on the resulting scene.

\subsection{An Error Report Improves Revision}
\label{sec:report-test}

\begin{table}[t]
  \centering
  \caption{\textbf{Revision gain with and without an error report.}
  Mean first-to-final gains in $G$ with estimated points, using one
  fixed registration per room. Both arms run the same check after each
  save; only report visibility differs.}
  \label{tab:report-main}
  \small
  \setlength{\tabcolsep}{4pt}
  \begin{tabular}{@{}lccc@{}}
    \toprule
    & \multicolumn{2}{c}{\textbf{Gain in $\boldsymbol{G}$}} & \\
    \cmidrule(lr){2-3}
    \textbf{Model} & \textbf{Hidden} & \textbf{Shown} & \textbf{Difference [95\% CI]} \\
    \midrule
    GPT-6.1 Terra & \WIRepTwoCtlGainMean{} & \WIRepTwoGainMean{} & \textbf{\WIRepTwoDiDMean{}} [\WIRepTwoDiDLo{}, \WIRepTwoDiDHi{}] \\
    GPT-6.1 Sol & \XRSolCtlGain{} & \XRSolGain{} & \textbf{\XRSolDiD{}} [\XRSolDiDLo{}, \XRSolDiDHi{}] \\
    GPT-6 Astra & \XRAstraCtlGain{} & \XRAstraGain{} & \XRAstraDiD{} [\XRAstraDiDLo{}, \XRAstraDiDHi{}] \\
    \bottomrule
  \end{tabular}
\end{table}

\begin{figure*}[t]
  \centering
  \includegraphics[width=\textwidth]{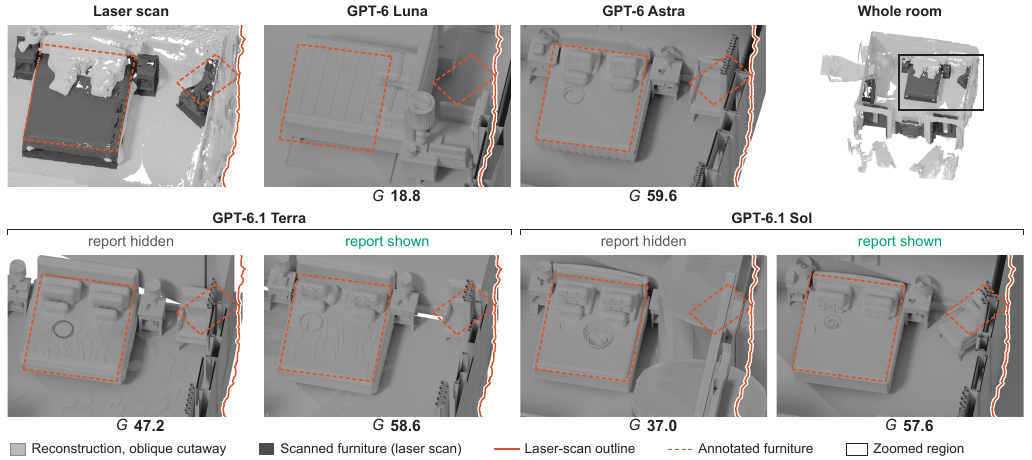}
  \caption{\textbf{Reconstructions with and without the error report.} Final scenes built from estimated points in one bedroom, cropped to the bed and adjacent chair; solid lines mark the laser scan and dashed boxes the annotated furniture, and $G$ is for the whole room. Top: the laser scan, GPT-6 Luna, GPT-6 Astra with the report hidden, and the whole room with the crop framed. Bottom: GPT-6.1 Terra and GPT-6.1 Sol with the report hidden and shown. Luna, shown at its best-matching rigid orientation, places the bed against the wrong wall.}
  \label{fig:report-examples}
\end{figure*}

We test whether reporting scene-to-input discrepancies improves
revision in the estimated-point condition, for Terra, Sol, and Astra.
After each save, the harness casts rays from the supplied cameras into
the scene to compare its visible surfaces with the estimated points.
It reports the fraction of visible surface within 10\,cm of these
points and lists up to five objects with the largest discrepancies,
giving a direction and distance for each.
The agent already has everything needed to compute this comparison,
so the report adds no geometric evidence.
It checks agreement with the supplied geometry, not whether that
geometry is itself accurate.

As a control, we run each agent again on the same rooms at the same
time. The harness performs the same check but does not show the
result.
Table~\ref{tab:report-main} shows that the report increases the
first-to-final gain in $G$ by \WIRepTwoDiDMean{} points for Terra and
\XRSolDiD{} points for Sol at equal spend, with 95\% confidence
intervals that exclude zero.
For Terra, the gain is \WRRigDiDMean{} points under our
standard rigid registration, about a third of the
\WRRigGapCtl{}-point gap between the control and the dense reference.

Figure~\ref{fig:report-examples} shows reconstructions with and
without the error report in one room.
For Astra, the interval includes zero.
Its revision already improves $G$ when the report is hidden, and its
final scenes with and without the report differ by only
\XRAstraFinalDiff{} points; the larger difference in gains reflects
lower first saves in the runs with the report.
Sol and Astra each corrected a grossly misplaced first scene in one
room after receiving the report.
Excluding these rooms reduces the differences from \XRSolDiD{} to
\XRSolDiDNoOutlier{} points for Sol and from \XRAstraDiD{} to
\XRAstraDiDNoOutlier{} points for Astra.

All three agents act on the report.
Among report lines followed by another save, the named object is
edited in \WRTouchedShownPct{}\%, \XRSolTouchedShown{}\%, and
\XRAstraTouchedShown{}\% of cases for Terra, Sol, and Astra when the
report is shown, and in \WRTouchedCtlPct{}\%, \XRSolTouchedCtl{}\%,
and \XRAstraTouchedCtl{}\% when it is hidden.
The report experiment shows that geometric evidence can support
correction as well as construction when it is used to check the
scene, for agents that do not make this check effective on their own.

%% file: fig/data/section7/section7_macros.tex
\newcommand{\WLRunsN}{1{,}040}
\newcommand{\WLStepsN}{54{,}842}
\newcommand{\WLToolCallsN}{53{,}793}
\newcommand{\WLSolFirstSaveVideoOnly}{6}
\newcommand{\WLSolFirstSavePoints}{14.5}
\newcommand{\WLTerraFirstSaveVideoOnly}{4}
\newcommand{\WLTerraFirstSavePoints}{14}
\newcommand{\WLAstraFirstSaveVideoOnly}{8}
\newcommand{\WLAstraFirstSavePoints}{23.5}
\newcommand{\WLSolTurnsVideoOnly}{10}
\newcommand{\WLSolTurnsVideoImage}{20}
\newcommand{\WLAstraTurnsVideoOnly}{26}
\newcommand{\WLAstraTurnsVideoImage}{46.5}
\newcommand{\WLTerraTurnsVideoOnly}{32.5}
\newcommand{\WLTerraTurnsVideoImage}{191}
\newcommand{\WLMeasureShareEvidenceMin}{18}
\newcommand{\WLMeasureShareEvidenceMax}{44}
\newcommand{\WLMeasureSharePointsMin}{29}
\newcommand{\WLMeasureSharePointsMax}{40}
\newcommand{\WLPointsGainMin}{22}
\newcommand{\WLPointsGainMax}{24}
\newcommand{\WLTerraMeasuredOwnRenderPct}{93}
\newcommand{\WLSolMeasuredOwnRenderPct}{78}
\newcommand{\WLAstraMeasuredOwnRenderPct}{53}
\newcommand{\WLAstraMeasuredFramePct}{26}
\newcommand{\WLSolVideoImageViewsN}{389}
\newcommand{\WLSolVideoImageOwnRenderN}{365}
\newcommand{\WLSolVideoImageFrameN}{0}
\newcommand{\WLSolVideoOnlyObjectSetRevisions}{0.8}
\newcommand{\WLSolVideoOnlyMeshObjectsAdded}{175}
\newcommand{\WLSolVideoOnlyMoveOnlyRevisions}{0.8}
\newcommand{\WLSolVideoImageObjectSetRevisions}{2.5}
\newcommand{\WLSolVideoImageMeshObjectsAdded}{1{,}090}
\newcommand{\WLSolVideoImageMoveOnlyRevisions}{1.4}
\newcommand{\WLAstraVideoOnlyObjectSetRevisions}{0.7}
\newcommand{\WLAstraVideoOnlyMeshObjectsAdded}{68}
\newcommand{\WLAstraVideoOnlyMoveOnlyRevisions}{1.5}
\newcommand{\WLAstraVideoImageObjectSetRevisions}{4.1}
\newcommand{\WLAstraVideoImageMeshObjectsAdded}{3{,}172}
\newcommand{\WLAstraVideoImageMoveOnlyRevisions}{1.9}
\newcommand{\WLAddRevisionMean}{1.2}
\newcommand{\WLAddRevisionLo}{0.3}
\newcommand{\WLAddRevisionHi}{2.2}
\newcommand{\WLAddRevisionN}{221}
\newcommand{\WLMoveRevisionMean}{0.1}
\newcommand{\WLMoveRevisionLo}{$-$0.6}
\newcommand{\WLMoveRevisionHi}{1.0}
\newcommand{\WLMoveRevisionN}{60}
\newcommand{\WLPairsTerraShare}{53}
\newcommand{\WLTerraCompositeRuns}{1}
\newcommand{\WLSolCompositeRuns}{5}
\newcommand{\WLAstraCompositeRuns}{13}
\newcommand{\WLSolCompositeDiffMean}{$-$0.3}
\newcommand{\WLSolCompositeDiffLo}{$-$8.3}
\newcommand{\WLSolCompositeDiffHi}{8.2}
\newcommand{\WLAstraCompositeDiffMean}{$-$1.5}
\newcommand{\WLAstraCompositeDiffLo}{$-$11.4}
\newcommand{\WLAstraCompositeDiffHi}{7.3}
\newcommand{\WLTracedPlacementsN}{190}
\newcommand{\WLTracedRunsN}{19}
\newcommand{\WLTracedDiffMedianCm}{2.5}
\newcommand{\WLLunaFurnitureUnmeasuredN}{24}
\newcommand{\WLLunaFurnitureTracedN}{28}
\newcommand{\WLSolPointsRemeasureRuns}{28}
\newcommand{\WLTerraPointsRemeasureRuns}{27}
\newcommand{\WLAstraPointsRemeasureRuns}{25}
\newcommand{\WLPointRunsN}{380}
\newcommand{\WLSceneVsPointsRuns}{11}
\newcommand{\WLStrongPreSaveShareMin}{51}
\newcommand{\WLStrongPreSaveShareMax}{59}
\newcommand{\WLStrongFirstObjectsMin}{1{,}800}
\newcommand{\WLStrongFirstObjectsMax}{3{,}200}
\newcommand{\WLMidPreSaveShareMin}{29}
\newcommand{\WLMidPreSaveShareMax}{34}
\newcommand{\WLMidFirstObjectsMin}{410}
\newcommand{\WLMidFirstObjectsMax}{890}
\newcommand{\WLLunaFirstObjects}{470}
\newcommand{\WLLunaPreMeasureSteps}{3}
\newcommand{\WLOthersPreMeasureStepsMin}{8}
\newcommand{\WLOthersPreMeasureStepsMax}{13.5}
\newcommand{\WLSolCheckSteps}{1}
\newcommand{\WLTerraCheckSteps}{13}
\newcommand{\WLDoneRunsN}{1{,}040}
\newcommand{\WLHalfUnspentN}{925}
\newcommand{\WLUncheckedFinalSaveN}{639}
\newcommand{\WLBestFrameVisiblePct}{81}
\newcommand{\WLAwayVisibleTwoThree}{40}
\newcommand{\WLAwayVisibleThreeFour}{12}
\newcommand{\WLAwayVisibleFourPlus}{0}
\newcommand{\WLBackPureVisiblePct}{23}
\newcommand{\WLDispObjectsN}{271}
\newcommand{\WLDispPairsN}{3{,}041}
\newcommand{\WLLateralVisiblePct}{59}
\newcommand{\WLAlongScaledVisiblePct}{23}
\newcommand{\WLPoseOffsetPxMin}{10}
\newcommand{\WLPoseOffsetPxMax}{12}
\newcommand{\WLFloorContactPerfectCm}{5}
\newcommand{\WLTriangulationPerfectCm}{22}
\newcommand{\WLTriangulationParallaxCm}{6}
\newcommand{\WLCameraReadRuns}{32}
\newcommand{\WLCameraReadCompared}{26}
\newcommand{\WLPointReadRuns}{24}
\newcommand{\WLPointReadCompared}{1}

%% file: fig/data/section7/section7_trace_addendum_macros.tex
\renewcommand{\WLRunsN}{1{,}060}
\renewcommand{\WLStepsN}{55{,}892}
\renewcommand{\WLToolCallsN}{54{,}823}
\renewcommand{\WLDoneRunsN}{1{,}060}
\renewcommand{\WLHalfUnspentN}{945}
\renewcommand{\WLAstraTurnsVideoImage}{45.5}
\renewcommand{\WLStrongPreSaveShareMin}{49}
\renewcommand{\WLStrongPreSaveShareMax}{59}
\renewcommand{\WLStrongFirstObjectsMin}{1{,}800}

%% file: fig/data/section7/section7_firstfinal_macros.tex
\newcommand{\WFTerraVideoRigidMean}{$-$0.2}
\newcommand{\WFTerraVideoRigidLo}{$-$2.4}
\newcommand{\WFTerraVideoRigidHi}{2.0}
\newcommand{\WFTerraVideoRigidN}{30}
\newcommand{\WFTerraVideoSimMean}{1.7}
\newcommand{\WFTerraVideoSimLo}{$-$1.8}
\newcommand{\WFTerraVideoSimHi}{5.1}
\newcommand{\WFTerraVideoSimN}{30}
\newcommand{\WFTerraVideoReplayOnlyRigidMean}{$-$0.6}
\newcommand{\WFTerraVideoReplayOnlyRigidN}{25}
\newcommand{\WFTerraVideoRuns}{30}
\newcommand{\WFTerraVideoMultiN}{30}
\newcommand{\WFTerraVideoMultiWithFirstN}{30}
\newcommand{\WFTerraPointsRigidMean}{1.6}
\newcommand{\WFTerraPointsRigidLo}{0.5}
\newcommand{\WFTerraPointsRigidHi}{3.0}
\newcommand{\WFTerraPointsRigidN}{30}
\newcommand{\WFTerraPointsSimMean}{1.7}
\newcommand{\WFTerraPointsSimLo}{0.7}
\newcommand{\WFTerraPointsSimHi}{3.0}
\newcommand{\WFTerraPointsSimN}{30}
\newcommand{\WFTerraPointsReplayOnlyRigidMean}{$-$0.1}
\newcommand{\WFTerraPointsReplayOnlyRigidN}{9}
\newcommand{\WFTerraPointsRuns}{30}
\newcommand{\WFTerraPointsMultiN}{30}
\newcommand{\WFTerraPointsMultiWithFirstN}{30}
\newcommand{\WFSolVideoRigidMean}{1.2}
\newcommand{\WFSolVideoRigidLo}{$-$1.5}
\newcommand{\WFSolVideoRigidHi}{3.9}
\newcommand{\WFSolVideoRigidN}{28}
\newcommand{\WFSolVideoSimMean}{3.5}
\newcommand{\WFSolVideoSimLo}{0.0}
\newcommand{\WFSolVideoSimHi}{7.7}
\newcommand{\WFSolVideoSimN}{28}
\newcommand{\WFSolVideoReplayOnlyRigidMean}{1.9}
\newcommand{\WFSolVideoReplayOnlyRigidN}{22}
\newcommand{\WFSolVideoRuns}{30}
\newcommand{\WFSolVideoMultiN}{28}
\newcommand{\WFSolVideoMultiWithFirstN}{28}
\newcommand{\WFSolPointsRigidMean}{0.6}
\newcommand{\WFSolPointsRigidLo}{$-$0.7}
\newcommand{\WFSolPointsRigidHi}{1.7}
\newcommand{\WFSolPointsRigidN}{29}
\newcommand{\WFSolPointsSimMean}{1.5}
\newcommand{\WFSolPointsSimLo}{0.7}
\newcommand{\WFSolPointsSimHi}{2.3}
\newcommand{\WFSolPointsSimN}{29}
\newcommand{\WFSolPointsReplayOnlyRigidMean}{0.4}
\newcommand{\WFSolPointsReplayOnlyRigidN}{15}
\newcommand{\WFSolPointsRuns}{30}
\newcommand{\WFSolPointsMultiN}{29}
\newcommand{\WFSolPointsMultiWithFirstN}{29}
\newcommand{\WFAstraVideoRigidMean}{$-$0.1}
\newcommand{\WFAstraVideoRigidLo}{$-$1.1}
\newcommand{\WFAstraVideoRigidHi}{0.9}
\newcommand{\WFAstraVideoRigidN}{27}
\newcommand{\WFAstraVideoSimMean}{0.2}
\newcommand{\WFAstraVideoSimLo}{$-$1.3}
\newcommand{\WFAstraVideoSimHi}{1.8}
\newcommand{\WFAstraVideoSimN}{27}
\newcommand{\WFAstraVideoReplayOnlyRigidMean}{$-$0.5}
\newcommand{\WFAstraVideoReplayOnlyRigidN}{13}
\newcommand{\WFAstraVideoRuns}{30}
\newcommand{\WFAstraVideoMultiN}{30}
\newcommand{\WFAstraVideoMultiWithFirstN}{27}
\newcommand{\WFAstraPointsRigidMean}{6.0}
\newcommand{\WFAstraPointsRigidLo}{$-$0.0}
\newcommand{\WFAstraPointsRigidHi}{13.7}
\newcommand{\WFAstraPointsRigidN}{29}
\newcommand{\WFAstraPointsSimMean}{7.9}
\newcommand{\WFAstraPointsSimLo}{2.0}
\newcommand{\WFAstraPointsSimHi}{15.6}
\newcommand{\WFAstraPointsSimN}{29}
\newcommand{\WFAstraPointsReplayOnlyRigidMean}{9.5}
\newcommand{\WFAstraPointsReplayOnlyRigidN}{18}
\newcommand{\WFAstraPointsRuns}{30}
\newcommand{\WFAstraPointsMultiN}{30}
\newcommand{\WFAstraPointsMultiWithFirstN}{29}
\newcommand{\WFLunaVideoRigidMean}{0.2}
\newcommand{\WFLunaVideoRigidLo}{$-$0.9}
\newcommand{\WFLunaVideoRigidHi}{1.6}
\newcommand{\WFLunaVideoRigidN}{14}
\newcommand{\WFLunaVideoSimMean}{$-$0.5}
\newcommand{\WFLunaVideoSimLo}{$-$0.9}
\newcommand{\WFLunaVideoSimHi}{$-$0.1}
\newcommand{\WFLunaVideoSimN}{14}
\newcommand{\WFLunaVideoReplayOnlyRigidMean}{0.2}
\newcommand{\WFLunaVideoReplayOnlyRigidN}{14}
\newcommand{\WFLunaVideoRuns}{30}
\newcommand{\WFLunaVideoMultiN}{14}
\newcommand{\WFLunaVideoMultiWithFirstN}{14}
\newcommand{\WFLunaPointsRigidMean}{$-$0.3}
\newcommand{\WFLunaPointsRigidLo}{$-$1.8}
\newcommand{\WFLunaPointsRigidHi}{1.0}
\newcommand{\WFLunaPointsRigidN}{24}
\newcommand{\WFLunaPointsSimMean}{$-$0.5}
\newcommand{\WFLunaPointsSimLo}{$-$2.7}
\newcommand{\WFLunaPointsSimHi}{1.4}
\newcommand{\WFLunaPointsSimN}{24}
\newcommand{\WFLunaPointsReplayOnlyRigidMean}{$-$0.3}
\newcommand{\WFLunaPointsReplayOnlyRigidN}{24}
\newcommand{\WFLunaPointsRuns}{30}
\newcommand{\WFLunaPointsMultiN}{25}
\newcommand{\WFLunaPointsMultiWithFirstN}{24}
\newcommand{\WFTerraVideoViewRigidMean}{$-$0.9}
\newcommand{\WFTerraVideoViewRigidLo}{$-$5.4}
\newcommand{\WFTerraVideoViewRigidHi}{2.8}
\newcommand{\WFTerraVideoViewRigidN}{30}
\newcommand{\WFTerraVideoViewSimMean}{$-$1.5}
\newcommand{\WFTerraVideoViewSimLo}{$-$6.1}
\newcommand{\WFTerraVideoViewSimHi}{2.4}
\newcommand{\WFTerraVideoViewSimN}{30}
\newcommand{\WFTerraVideoViewReplayOnlyRigidMean}{1.1}
\newcommand{\WFTerraVideoViewReplayOnlyRigidN}{23}
\newcommand{\WFTerraVideoViewRuns}{30}
\newcommand{\WFTerraVideoViewMultiN}{30}
\newcommand{\WFTerraVideoViewMultiWithFirstN}{30}
\newcommand{\WFSolVideoViewRigidMean}{1.0}
\newcommand{\WFSolVideoViewRigidLo}{$-$1.5}
\newcommand{\WFSolVideoViewRigidHi}{3.7}
\newcommand{\WFSolVideoViewRigidN}{30}
\newcommand{\WFSolVideoViewSimMean}{2.1}
\newcommand{\WFSolVideoViewSimLo}{$-$0.9}
\newcommand{\WFSolVideoViewSimHi}{5.3}
\newcommand{\WFSolVideoViewSimN}{30}
\newcommand{\WFSolVideoViewReplayOnlyRigidMean}{1.1}
\newcommand{\WFSolVideoViewReplayOnlyRigidN}{20}
\newcommand{\WFSolVideoViewRuns}{30}
\newcommand{\WFSolVideoViewMultiN}{30}
\newcommand{\WFSolVideoViewMultiWithFirstN}{30}
\newcommand{\WFLunaVideoViewRigidMean}{$-$0.6}
\newcommand{\WFLunaVideoViewRigidLo}{$-$1.8}
\newcommand{\WFLunaVideoViewRigidHi}{0.5}
\newcommand{\WFLunaVideoViewRigidN}{28}
\newcommand{\WFLunaVideoViewSimMean}{0.5}
\newcommand{\WFLunaVideoViewSimLo}{$-$1.1}
\newcommand{\WFLunaVideoViewSimHi}{2.2}
\newcommand{\WFLunaVideoViewSimN}{28}
\newcommand{\WFLunaVideoViewReplayOnlyRigidMean}{$-$0.6}
\newcommand{\WFLunaVideoViewReplayOnlyRigidN}{28}
\newcommand{\WFLunaVideoViewRuns}{30}
\newcommand{\WFLunaVideoViewMultiN}{30}
\newcommand{\WFLunaVideoViewMultiWithFirstN}{28}
\newcommand{\WFAstraVideoViewRigidMean}{3.6}
\newcommand{\WFAstraVideoViewRigidLo}{0.9}
\newcommand{\WFAstraVideoViewRigidHi}{6.6}
\newcommand{\WFAstraVideoViewRigidN}{29}
\newcommand{\WFAstraVideoViewSimMean}{4.0}
\newcommand{\WFAstraVideoViewSimLo}{0.8}
\newcommand{\WFAstraVideoViewSimHi}{7.4}
\newcommand{\WFAstraVideoViewSimN}{29}
\newcommand{\WFAstraVideoViewReplayOnlyRigidMean}{4.5}
\newcommand{\WFAstraVideoViewReplayOnlyRigidN}{12}
\newcommand{\WFAstraVideoViewRuns}{30}
\newcommand{\WFAstraVideoViewMultiN}{30}
\newcommand{\WFAstraVideoViewMultiWithFirstN}{29}
\newcommand{\WFGrokVideoViewRigidMean}{$-$0.4}
\newcommand{\WFGrokVideoViewRigidLo}{$-$2.6}
\newcommand{\WFGrokVideoViewRigidHi}{1.5}
\newcommand{\WFGrokVideoViewRigidN}{10}
\newcommand{\WFGrokVideoViewSimMean}{$-$2.3}
\newcommand{\WFGrokVideoViewSimLo}{$-$5.9}
\newcommand{\WFGrokVideoViewSimHi}{1.0}
\newcommand{\WFGrokVideoViewSimN}{10}
\newcommand{\WFGrokVideoViewRuns}{10}
\newcommand{\WFGrokVideoViewMultiN}{10}
\newcommand{\WFGrokVideoViewMultiWithFirstN}{10}
\newcommand{\WFReplayReproducedN}{223}
\newcommand{\WFMeasTerraCameraDirectMean}{7.7}
\newcommand{\WFMeasTerraCameraDirectLo}{4.0}
\newcommand{\WFMeasTerraCameraDirectHi}{11.8}
\newcommand{\WFMeasTerraCameraRigidMean}{4.7}
\newcommand{\WFMeasTerraCameraRigidLo}{1.4}
\newcommand{\WFMeasTerraCameraRigidHi}{8.0}
\newcommand{\WFMeasTerraCameraSimMean}{4.6}
\newcommand{\WFMeasTerraCameraSimLo}{0.9}
\newcommand{\WFMeasTerraCameraSimHi}{8.2}
\newcommand{\WFMeasTerraCameraN}{23}
\newcommand{\WFMeasTerraCameraImageDirectMean}{7.0}
\newcommand{\WFMeasTerraCameraImageDirectLo}{1.6}
\newcommand{\WFMeasTerraCameraImageDirectHi}{12.9}
\newcommand{\WFMeasTerraCameraImageRigidMean}{3.2}
\newcommand{\WFMeasTerraCameraImageRigidLo}{$-$1.4}
\newcommand{\WFMeasTerraCameraImageRigidHi}{7.5}
\newcommand{\WFMeasTerraCameraImageSimMean}{2.3}
\newcommand{\WFMeasTerraCameraImageSimLo}{$-$2.2}
\newcommand{\WFMeasTerraCameraImageSimHi}{6.5}
\newcommand{\WFMeasTerraCameraImageN}{12}
\newcommand{\WFMeasTerraCameraImageTenDirectMean}{13.0}
\newcommand{\WFMeasTerraCameraImageTenDirectLo}{7.6}
\newcommand{\WFMeasTerraCameraImageTenDirectHi}{19.5}
\newcommand{\WFMeasTerraCameraImageTenRigidMean}{5.6}
\newcommand{\WFMeasTerraCameraImageTenRigidLo}{2.1}
\newcommand{\WFMeasTerraCameraImageTenRigidHi}{9.0}
\newcommand{\WFMeasTerraCameraImageTenSimMean}{5.1}
\newcommand{\WFMeasTerraCameraImageTenSimLo}{1.0}
\newcommand{\WFMeasTerraCameraImageTenSimHi}{9.5}
\newcommand{\WFMeasTerraCameraImageTenN}{17}
\newcommand{\WFMeasSolCameraDirectMean}{3.6}
\newcommand{\WFMeasSolCameraDirectLo}{1.4}
\newcommand{\WFMeasSolCameraDirectHi}{6.1}
\newcommand{\WFMeasSolCameraRigidMean}{3.9}
\newcommand{\WFMeasSolCameraRigidLo}{0.8}
\newcommand{\WFMeasSolCameraRigidHi}{7.5}
\newcommand{\WFMeasSolCameraSimMean}{3.4}
\newcommand{\WFMeasSolCameraSimLo}{$-$0.3}
\newcommand{\WFMeasSolCameraSimHi}{7.4}
\newcommand{\WFMeasSolCameraN}{22}
\newcommand{\WFMeasSolCameraImageDirectMean}{4.3}
\newcommand{\WFMeasSolCameraImageDirectLo}{1.3}
\newcommand{\WFMeasSolCameraImageDirectHi}{7.2}
\newcommand{\WFMeasSolCameraImageRigidMean}{2.0}
\newcommand{\WFMeasSolCameraImageRigidLo}{$-$1.1}
\newcommand{\WFMeasSolCameraImageRigidHi}{5.4}
\newcommand{\WFMeasSolCameraImageSimMean}{0.9}
\newcommand{\WFMeasSolCameraImageSimLo}{$-$3.2}
\newcommand{\WFMeasSolCameraImageSimHi}{5.1}
\newcommand{\WFMeasSolCameraImageN}{19}
\newcommand{\WFMeasAstraCameraDirectMean}{4.7}
\newcommand{\WFMeasAstraCameraDirectLo}{2.3}
\newcommand{\WFMeasAstraCameraDirectHi}{7.4}
\newcommand{\WFMeasAstraCameraRigidMean}{2.4}
\newcommand{\WFMeasAstraCameraRigidLo}{0.2}
\newcommand{\WFMeasAstraCameraRigidHi}{4.8}
\newcommand{\WFMeasAstraCameraSimMean}{3.4}
\newcommand{\WFMeasAstraCameraSimLo}{1.8}
\newcommand{\WFMeasAstraCameraSimHi}{5.2}
\newcommand{\WFMeasAstraCameraN}{11}
\newcommand{\WFMeasAstraCameraImageDirectMean}{6.6}
\newcommand{\WFMeasAstraCameraImageDirectLo}{3.7}
\newcommand{\WFMeasAstraCameraImageDirectHi}{10.1}
\newcommand{\WFMeasAstraCameraImageRigidMean}{2.7}
\newcommand{\WFMeasAstraCameraImageRigidLo}{$-$1.2}
\newcommand{\WFMeasAstraCameraImageRigidHi}{6.3}
\newcommand{\WFMeasAstraCameraImageSimMean}{2.3}
\newcommand{\WFMeasAstraCameraImageSimLo}{$-$0.2}
\newcommand{\WFMeasAstraCameraImageSimHi}{4.9}
\newcommand{\WFMeasAstraCameraImageN}{18}
\newcommand{\WFMeasAllCameraDirectMean}{5.5}
\newcommand{\WFMeasAllCameraDirectLo}{3.6}
\newcommand{\WFMeasAllCameraDirectHi}{7.6}
\newcommand{\WFMeasAllCameraRigidMean}{3.9}
\newcommand{\WFMeasAllCameraRigidLo}{2.0}
\newcommand{\WFMeasAllCameraRigidHi}{6.0}
\newcommand{\WFMeasAllCameraSimMean}{3.9}
\newcommand{\WFMeasAllCameraSimLo}{1.7}
\newcommand{\WFMeasAllCameraSimHi}{6.1}
\newcommand{\WFMeasAllCameraN}{56}
\newcommand{\WFMeasAllCameraImageDirectMean}{5.8}
\newcommand{\WFMeasAllCameraImageDirectLo}{3.6}
\newcommand{\WFMeasAllCameraImageDirectHi}{8.1}
\newcommand{\WFMeasAllCameraImageRigidMean}{2.6}
\newcommand{\WFMeasAllCameraImageRigidLo}{0.4}
\newcommand{\WFMeasAllCameraImageRigidHi}{4.8}
\newcommand{\WFMeasAllCameraImageSimMean}{1.8}
\newcommand{\WFMeasAllCameraImageSimLo}{$-$0.4}
\newcommand{\WFMeasAllCameraImageSimHi}{3.9}
\newcommand{\WFMeasAllCameraImageN}{49}
\newcommand{\WFMeasAllPooledDirectMean}{6.7}
\newcommand{\WFMeasAllPooledDirectLo}{5.1}
\newcommand{\WFMeasAllPooledDirectHi}{8.3}
\newcommand{\WFMeasAllPooledRigidMean}{3.6}
\newcommand{\WFMeasAllPooledRigidLo}{2.3}
\newcommand{\WFMeasAllPooledRigidHi}{5.0}
\newcommand{\WFMeasAllPooledSimMean}{3.2}
\newcommand{\WFMeasAllPooledSimLo}{1.7}
\newcommand{\WFMeasAllPooledSimHi}{4.7}
\newcommand{\WFMeasAllPooledN}{122}

%% file: fig/appearance_effects.tex
\begin{figure}[t]
\centering
\IfFileExists{fig/appearance_geometry_effects_v3.pdf}{%
  \includegraphics[width=0.72\linewidth]{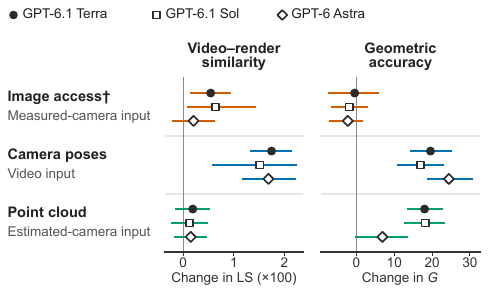}%
}{%
  \fbox{\begin{minipage}[c][1.25in][c]{0.93\linewidth}\centering\small
    \textit{Author review: artwork pending.}\\[5pt]
    \textbf{Former Figure 1b, moved here}\\[5pt]
    Video--render similarity change ($\Delta$LS$\times100$)\\
    Geometric accuracy change ($\Delta G$)\\[4pt]
    Enable image viewing / add cameras / add points
  \end{minipage}}%
}
\caption{\textbf{Changes in render similarity do not consistently reflect geometric gains.}
Paired changes when image access is enabled with measured-camera input$^\dagger$, measured camera poses are added to video input, and estimated points are added to estimated-camera input. LS is video--render latent similarity and $G$ is geometric accuracy. Intervals are 95\% scene-bootstrap intervals. $^\dagger$Image access uses the supplied-frame score $G_{\mathrm{direct}}$; the other contrasts use rigidly registered $G$.}
\label{fig:appearance-effects}
\end{figure}

%% file: sec/6_discussion.tex
\section{Limitations}
\label{sec:limitations}

Our study covers \PaperSceneN{} indoor rooms from one dataset, three
main models from one developer, and one harness, with a single run
per room and condition.
The error report was designed on the rooms on which it was tested,
so its transfer to other rooms and to agents from other developers is
untested.
It checks agreement with the supplied points, not their accuracy;
whether independent metric cues let agents correct inherited scale
errors remains open.
We could not confirm that our rooms were held out of DA3's training.
$G$ covers visible geometry only and does not measure object
semantics or how editable the programs are.
\section{Conclusion}
\label{sec:conclusion}

We introduced scan-referenced evaluation and geometric error reporting
to measure and improve scene geometry.
Video-only reconstructions fall well below dense references from the
same video, and render similarity
offers little guidance on geometric accuracy in individual rooms.
Given estimated poses and points, the strongest tested agent nearly
matches the dense reference's geometric accuracy in an editable scene,
but inherits the supplied scale errors.
Viewing renders improves appearance more consistently than geometry,
and execution traces show that agents measure their inputs but rarely
check their scenes against them.
Our error report makes this comparison explicit, improving geometric
accuracy during revision for GPT-6.1 Terra and GPT-6.1 Sol without
adding geometric evidence.
Geometric verification should therefore be an explicit part of the
reconstruction loop, allowing the same evidence to guide correction
as well as construction.